\documentclass{article} 
\usepackage{iclr2026_conference,times}

\usepackage{amsmath,amsfonts,bm}

\def\eqref#1{equation~\ref{#1}}

\def\1{\bm{1}}

\DeclareMathAlphabet{\mathsfit}{\encodingdefault}{\sfdefault}{m}{sl}
\SetMathAlphabet{\mathsfit}{bold}{\encodingdefault}{\sfdefault}{bx}{n}

\usepackage{hyperref}
\usepackage{url}
\usepackage{graphicx}
\usepackage{enumitem}
\usepackage{booktabs}
\usepackage{amssymb}
\usepackage{multirow}
\usepackage{subcaption}

\usepackage{pythonhighlight}
\usepackage{fancyvrb}
\usepackage{listings}
\usepackage[most]{tcolorbox}
\tcbuselibrary{listingsutf8}

\usepackage{wrapfig}

\title{Say One Thing, Do Another? Diagnosing Reasoning-Execution Gaps in VLM-Powered Mobile-Use Agents}

\author{Lingzhong Dong$^1$, Ziqi Zhou$^1$, Shuaibo Yang$^1$, Haiyue Sheng$^{2}$, Pengzhou Cheng$^1$,\\
\bf Zongru Wu$^1$, Zheng Wu$^1$, Gongshen Liu$^1$$^*$, Zhuosheng Zhang$^1$\thanks{Corresponding authors.}\\
$^1$Shanghai Jiao Tong University \quad $^2$Beijing Institute of Technology\\
\texttt{{\{lingzhong,lgshen,zhangzs\}}@sjtu.edu.cn}
}

\iclrfinalcopy 
\begin{document}

\maketitle

\begin{abstract}
Mobile-use agents powered by vision-language models (VLMs) have shown great potential in interpreting natural language instructions and generating corresponding actions based on mobile graphical user interface.
Recent studies suggest that incorporating chain-of-thought (CoT) reasoning tends to improve the execution accuracy.
However, existing evaluations emphasize execution accuracy while neglecting whether CoT reasoning aligns with ground-truth actions. 
This oversight fails to assess potential reasoning-execution gaps, which in turn foster over-trust: users relying on seemingly plausible CoTs may unknowingly authorize harmful actions, potentially resulting in financial loss or trust crisis.
In this work, we introduce a new evaluation framework to diagnose reasoning-execution gaps. At its core lies Ground-Truth Alignment (GTA), which measures whether the action implied by a CoT matches the ground-truth action. 
By combining GTA with the standard Exact Match (EM) metric, we jointly assess both the reasoning accuracy and execution accuracy. This joint perspective reveals two types of reasoning-execution gaps: (i) Execution Gap (EG), where the reasoning correctly identifies the correct action but execution fails, and (ii) Reasoning Gap (RG), where execution succeeds but reasoning process conflicts with the actual execution. 
Experimental results across a wide range of mobile interaction tasks reveal that reasoning-execution gaps are prevalent, with execution gaps occurring more frequently than reasoning gaps. Moreover, while scaling up model size reduces the overall gap, sizable execution gaps persist even in the largest models.
Further analysis shows that our framework reliably reflects systematic EG/RG patterns in state-of-the-art models.
These findings offer concrete diagnostics and support the development of more trustworthy mobile-use agents. Our data and code are publicly available at \href{https://github.com/LZ-Dong/Reasoning-Executing-Gaps}{~\texttt{https://github.com/LZ-Dong/Reasoning-Executing-Gaps}}.
\end{abstract}

\section{Introduction}

Mobile-use agents powered by vision-language models (VLMs) are increasingly capable of following natural language instructions and operating graphical user interfaces (GUIs) on mobile devices. By interpreting screenshots and mapping them to executable actions, these agents hold promise for tasks such as app navigation, automation, and accessibility support~\citep{wang2024gui,nguyen2024gui,zhang2024large,liu2025llm,hu2025agents}. Recent studies further show that incorporating chain-of-thought (CoT) reasoning can enhance execution accuracy, as intermediate reasoning steps help models decompose complex instructions and align with user intent~\citep{zhang2025agentcpm,qin2025ui,ye2025mobile,zhang2025does}.

Despite these advances, current evaluation practices remain limited. The dominant metric, Exact Match (EM), checks whether the predicted action exactly matches the ground truth~\citep{zhang2025agentcpm,wang2025mmbench,shi2025towards}. However, EM alone overlooks whether the CoT is consistent with the target action, obscuring cases where the agent acts correctly with the wrong reason or fails despite plausible reasoning. Such reasoning-execution gaps pose risks for over-trust, complicate debugging, and undermine the reliability of these systems~\citep{barez2025chain,zhao2025chain,chen2505reasoning,matton2025walk}.

This gap highlights the need for a principled evaluation framework that explicitly accounts for reasoning-execution gaps. A key challenge is to determine whether a CoT faithfully implies the ground-truth action and to disentangle whether errors arise from flawed reasoning or faulty execution. Without such diagnostics, existing metrics mix distinct error sources, offering limited insight into model behavior~\citep{shi2025towards,tang2025magicgui}.

To address this challenge, we introduce Ground-Truth Alignment (GTA), a new metric that evaluates whether the action implied by the CoT matches the ground-truth action. By combining GTA with the conventional EM metric, we derive a four-quadrant diagnostic framework that categorizes model outputs into: (i) Ideal, where both reasoning and action correct, (ii) Execution Gap (EG), where reasoning is correct but execution fails, (iii) Both Wrong, where both reasoning and action are incorrect, and (iv) Reasoning Gap (RG), where the action is correct but reasoning is inconsistent. This framework provides fine-grained insights into where and why models fail.

Our contributions can be summarized as follows:

(i) We introduce GTA, a principled metric that measures whether the action implied by an agent’s CoT align with the ground-truth action. By combining GTA with the standard EM metric, we derive a four-quadrant diagnostic space that disentangles reasoning accuracy from execution accuracy and reveals distinct error modes.

(ii) We design an automatic GTA Evaluator that maps free-form CoT reasoning into GTA score. Through a stratified sampled human annotation study, we validate that the evaluator provides reliable and reproducible assessments, making large-scale reasoning diagnostics feasible without costly manual labeling.

(iii) We conduct extensive experiments on diverse mobile-interaction benchmarks, including AITZ~\citep{zhang2024android}, CAGUI~\citep{zhang2025agentcpm}, and AndroidControl~\citep{li2024effects}, systematically quantify and characterize two key failure modes: EG and RG. Our results show that reasoning-execution gaps are widespread and that EG occurs more frequently, even in the largest state-of-the-art VLM agents.
\section{Related Work}
In this section, we first review the development of mobile-use agents and discuss the recent progress in leveraging CoT reasoning to improve task performance. Then, we dive into the studies of the faithfulness of mobile-use agents that forms the basis of this work.

\subsection{Mobile-Use Agents}
Mobile-use agents aim to operate smartphone applications autonomously, perceiving dynamic GUI states and performing fine-grained actions such as taps, swipes, and text input. Recent advances in VLMs~\citep{wang2024qwen2,bai2025qwen2,yao2024minicpm,openai2024gpt4o,openai2025gpt5,comanici2025gemini,team2024gemini} have enabled agents to process screenshots and natural language instructions directly. These developments have sparked intensive research on how to improve reasoning~\citep{zhang2025agentcpm,ye2025mobile,zhang2025btl}, grounding~\citep{wu2024atlas,cheng2024seeclick,zhou2025gui,gou2024navigating}, and reliability~\citep{cheng-etal-2025-os} in mobile-use agent. According to Mobile-Agent-v3~\citep{ye2025mobile}, existing mobile-use agents can be devided into two categories: single-agent and multi-agent system.

Single-agent methods usually follow an agent-as-a-model paradigm~\citep{chen2024spa}, where a single VLM is trained through continue pretraining (CPT), supervised fine-tuning (SFT), and reinforcement learning (RL) to jointly handle perception, reasoning, and action prediction. Early works (e.g., UGround~\citep{gou2024navigating}, OS-Atlas~\citep{wu2024atlas}, CogAgent~\citep{hong2024cogagent}, UI-TARS~\citep{qin2025ui}) demonstrate that combining GUI-specific pretraining with SFT on mobile interaction data yields competitive results. More recent research focuses on reinforcement fine-tuning (RFT) strategies, such as GRPO, to enhance reasoning ability. Representative RFT-based systems include UI-R1~\citep{lu2025ui}, GUI-R1~\citep{luo2025gui}, InfiGUI-R1~\citep{liu2025infigui}, AgentCPM-GUI~\citep{zhang2025agentcpm}, GUI-OWL~\citep{ye2025mobile}, MagicGUI~\citep{tang2025magicgui}, UI-Venus~\citep{gu2025ui}, Mobile-R1~\citep{gu2025mobile}, and BTL-UI~\citep{zhang2025btl}.

Multi-agent system decomposes mobile-use agents into specialized roles. For example, planner-executor frameworks~\citep{wang2024mobile, xu2025aguvis,guo2025atomic},separate high-level task decomposition from low-level interaction; memory-augmented systems~\citep{zhang2025appagent, wang2024agent,agashe2025agent} enhance long-horizon consistency; and reflection-based designs~\citep{li2025mobileuse, zheng2024gpt} iteratively refine decisions before execution.

Unlike prior work focused on execution accuracy, our study investigates whether an agent’s internal reasoning process faithfully supports its predicted actions. We introduce an evaluation framework to diagnose reasoning-execution gaps. This perspective provides a new lens to analyze mobile-use agents, complementing existing benchmarks and highlighting overlooked reliability issues.

\subsection{Mobile-Use Agents with CoT}
Given the proven success of Chain-of-Thought reasoning in large language models (LLMs)~\citep{zhang2025vitcot, xu2025mixed, cheng2025visual, cheng2025comt}, researchers have recently extended this paradigm to VLM-powered agents. By explicitly modeling intermediate reasoning steps, CoT enhances execution accuracy in complex tasks. In the domain of mobile-use agents, CoT has shown notable benefits: it boosts execution accuracy, strengthens human-agent interaction, and organizes historical context, thereby becoming central to modern GUI automation systems~\citep{zhang2025does, qin2025ui, ye2025mobile, liu2025infigui, luo2025gui}.

The CoT training paradigm evolved from SFT to RFT. Early systems integrated CoT into mobile-use agents through SFT~\citep{zhang2023you,zhang2024android,hong2024cogagent}. However, SFT depends on large amounts of expert-annotated data and shows weak robustness when applied to out-of-distribution tasks. To address these issues, recent work adopts RFT~\citep{zhang2025agentcpm,liu2025infigui,tang2025magicgui,luo2025gui,lu2025ui,zhang2025btl,gu2025mobile}, to better elicit reasoning ability of mobile-use agents.

Despite these advances, evidence shows that the current mobile-use agents struggles with reasoning-execution gaps~\citep{gu2025ui}. However, there is still a lack of studies on analyzing and understanding the challenge. 
This paper fills the gap by proposing a novel metric that disentangles reasoning-execution gaps into two distinct components: the execution gap and the reasoning gap. 
We conduct a systematic analysis of state-of-the-art models using this metric, offering new insights into their performance and failure modes.

\subsection{Faithfulness of Mobile-Use Agents}
With the increasing deployment of mobile-use agents across a wide range of real-world applications, a fundamental open question remains: can these agents remain faithful in complex, dynamic environments~\citep{zhang2024llamatouch,ma2025caution,shi2025towards}?
The challenge of faithfulness in such agents stems from two key dimensions: faithfulness to user intentions and faithfulness to the agent’s own decision-making process.

Regarding the first dimension, recent studies have shown that mobile agents are highly susceptible to environmental distractions, such as pop-up boxes~\citep{ma2025caution,zhang-etal-2025-attacking,chen2025aeia}. These distractions can derail agents from pursuing user-specified goals and may lead them into unintended or even unsafe environmental states.
Interestingly, applying CoT (e.g., instructing agents to pay attention to potential distractions) does not alleviate the issue, but even increases the likelihood of distraction. This  suggests that naive application of CoT may not inherently improve agent faithfulness.

For the second dimension, prior work has explored methods for measuring the confidence of decision making process~\citep{cheng-etal-2025-os,hao2025uncertainty,tao2025understanding}, thus allowing agents to proactively query users in uncertain or potentially harmful situations~\citep{wu2025verios,ai2025inquiremobile,cheng2025navi}. For example, VeriOS-Agent~\citep{wu2025verios} is capable of identifying untrustworthy scenarios, asking appropriate clarification queries, and following clarified instructions in such settings without degrading performance in routine tasks.

Despite these advances, there remains a gap in foundational understanding of reasoning-execution gaps, which underpins both aspects of agent faithfulness. 
This paper addresses this gap by introducing a new diagnostic framework with fine-grained metrics designed to systematically quantify the reasoning-execution gaps. Through this, we aim to deepen understanding of mobile-use agent behavior and support the development of more reliable agents suitable for real-world deployment.
\section{Method}
In this section, we first introduce the VLM-based mobile agent and formalize the task. We then present our evaluation metrics, Exact Match (EM) for execution accuracy and Ground-Truth Alignment (GTA) for reasoning accuracy, along with a diagnostic framework. Finally, we describe the design of the GTA Evaluator, a key component for assessing reasoning quality.
\subsection{VLM-based Mobile Agent}
This paper investigates VLM-based mobile agents, which generate CoT reasoning and executable actions directly from natural language instructions and GUI screenshots. 
Formally, the end-to-end task is defined as modeling the conditional probability:
\begin{equation}
P(c_n, a_n \mid I, H_n, o_n),
\end{equation}
where $I$ denotes the task instruction, $o_n$ represents the current screenshot, and $H_n$ denotes optional history context. 
The representation of $H_n$ varies across agents, depending on their design and training paradigms.
For example, AgentCPM-GUI~\citep{zhang2025agentcpm} does not incorporate history, thus treating $H_n$ as empty. 
In contrast, UI-TARS~\citep{qin2025ui} encodes $H_n$ with dialogue-style history format, while GUI-Owl~\citep{ye2025mobile} compresses the historical context for efficient integration. Detailed descriptions are provided in the Appendix~\ref{app:models}.
Importantly, we focus on models that explicitly generate $(c_n,a_n)$, which enables us to analyze the reasoning-execution gaps.

\subsection{Evaluation Metrics}
Following standard GUI benchmarks~\citep{zhang2025agentcpm,wang2025mmbench,shi2025towards}, we adopt \textbf{Exact Match (EM)} for execution accuracy, and further introduce \textbf{Ground-Truth Alignment (GTA)} for reasoning accuracy.

\paragraph{Exact Match (EM).}
Let $a_n^{*}$ be the ground-truth action at step $n$ and $a_n$ the model’s executed action. Formally, let $\mathbf{1}_{\{\cdot\}}$ denote the indicator function:
\begin{equation}
\mathrm{EM}_n = \mathbf{1}_{\{a_n = a_n^{*}\}},
\end{equation}
where equality requires both the action type and parameters to match. The overall EM score is defined as the step-level average:
\begin{equation}
\mathrm{EM} = \frac{1}{N}\sum_{n=1}^N \mathrm{EM}_n.
\end{equation}

\paragraph{Ground-Truth Alignment (GTA).}
Let $c_n$ denote the CoT generated at step $n$, and $f(c_n)$ the action implied by the CoT:
\begin{equation}
\mathrm{GTA}_n = \mathbf{1}_{\{f(c_n) = a_n^{*}\}},
\end{equation}
measuring whether the CoT correctly leads to the ground-truth action. The overall GTA score is
\begin{equation}
\mathrm{GTA} = \frac{1}{N}\sum_{n=1}^N \mathrm{GTA}_n.
\end{equation}

\begin{wrapfigure}{r}{0.48\linewidth} 
    \centering
    \vspace{-5mm}
    \includegraphics[width=\linewidth]{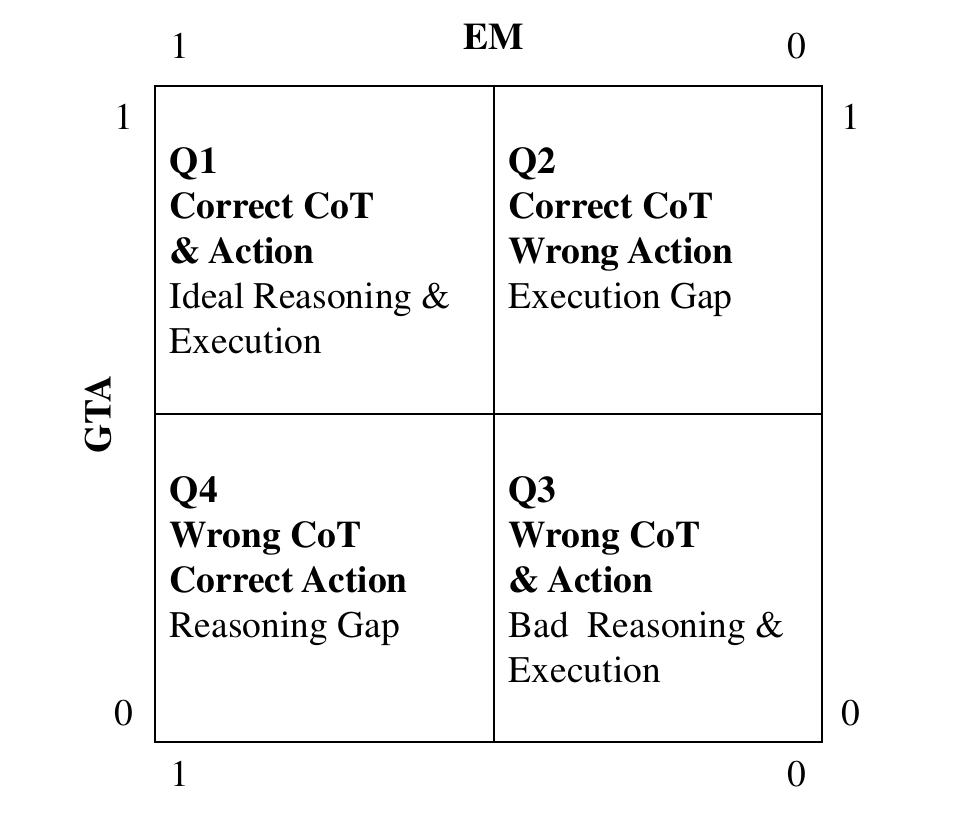}
    \vspace{-5mm}
    \caption{
    Four-quadrant diagnostic framework of reasoning–execution gaps. 
    The axes represent reasoning accuracy (GTA) and action accuracy (EM). 
    Q1:  Ideal, where both reasoning and action correct; Q2: Execution Gap (EG), where reasoning is correct but execution fails; Q3 Both Wrong, where both reasoning and action are incorrect;
    Q4: Reasoning Gap (RG), where the action is correct but reasoning fails.
    }
    \vspace{-8mm}
    \label{fig:quadrant}
\end{wrapfigure}

\paragraph{Quadrant Analysis.}
Combining EM and GTA yields a four-quadrant categorization (Figure~\ref{fig:quadrant}):
\begin{itemize}
    \item Q1: \(\mathrm{EM}_n=1, \mathrm{GTA}_n=1\) (Ideal, where both reasoning and action correct)
    \item Q2: \(\mathrm{EM}_n=0, \mathrm{GTA}_n=1\) (Execution Gap, where reasoning is correct but execution fails)
    \item Q3: \(\mathrm{EM}_n=0, \mathrm{GTA}_n=0\) (Both Wrong, where both reasoning and action are incorrect)
    \item Q4: \(\mathrm{EM}_n=1, \mathrm{GTA}_n=0\) (Reasoning Gap, where the action is correct but reasoning fails)
\end{itemize}

\paragraph{Diagnostic Rates.}
Beyond individual metrics, we further define two diagnostic rates to capture mismatch patterns between reasoning and action:
\begin{equation}
\mathrm{EG} = \frac{1}{N}\sum_{n=1}^N \mathbf{1}_{\{\mathrm{GTA}_n = 1 \wedge \mathrm{EM}_n = 0\}},
\end{equation}
\begin{equation}
\mathrm{RG} = \frac{1}{N}\sum_{n=1}^N \mathbf{1}_{\{\mathrm{GTA}_n = 0 \wedge \mathrm{EM}_n = 1\}}.
\end{equation}
Here, \textbf{Execution Gap (EG)} quantifies the fraction of steps where the CoT implies the correct action but the executed action is wrong (\emph{thought correct, action wrong}), while \textbf{Reasoning Gap (RG)} quantifies the fraction of steps where the executed action is correct but the CoT is wrong (\emph{thought wrong, action correct}).

\subsection{GTA Evaluator}
A key challenge in computing $\mathrm{GTA}$ lies in mapping the free-form CoT $c_n$ into its implied action $f(c_n)$. 
Since CoT is often expressed in natural language, we design an automatic evaluator based on an instruction-following VLM. 
Specifically, we formulate the mapping as
\begin{equation}
f(c_n) = \arg\max_{a} P(a \mid c_n, H_n, o_n),
\end{equation}
where the evaluator predicts the most likely executable action $a$ given the CoT reasoning $c_n$, the local history $H_n$, and the current observation $o_n$. 
We enforce deterministic decoding (greedy decoding) to guarantee reproducibility of $f(c_n)$.

We retain $(H_n, o_n)$ in the evaluator’s input because CoT $c_n$ often contains underspecified or context-dependent references 
(e.g., ``click the confirm button below’’). 
Without grounding in the current screenshot $o_n$ and dialogue history $H_n$, the implied action could be ambiguous or even infeasible. 
Conditioning on $(c_n, H_n, o_n)$ therefore ensures that $f(c_n)$ consistently maps reasoning traces to executable actions in the same action space as the ground-truth annotation.

Finally, the evaluation of $\mathrm{GTA}$ follows the same matching rule as $\mathrm{EM}$: 
both the action type and all parameters must match the ground-truth annotation. 
This consistent criterion ensures that reasoning accuracy (GTA) and execution accuracy (EM) are directly comparable.

\section{Experiment}
In this section, we conduct comprehensive experiments  to address the following research questions:
\begin{itemize}[leftmargin=*]
    \item \textbf{RQ1:} To what extent does the proposed GTA Evaluator provide reliable and reproducible measurements of reasoning accuracy? 
    \item \textbf{RQ2:} How do state-of-the-art VLM-based mobile agents perform in terms of reasoning accuracy, execution accuracy, and reasoning-execution gaps across public benchmarks?
    \item \textbf{RQ3:} What is the impact of parameter scaling on reasoning accuracy, execution accuracy, and the magnitude of reasoning-execution gaps?
\end{itemize}

\subsection{Experiment Setup}
We structure our experiments on three series of mobile-use agents and three challenging datasets to evaluate their reasoning accuracy (GTA) and execution accuracy (EM).

\paragraph{Models} For our experiments, we select three state-of-the-art mobile-use agents that represent the strongest open-source baselines across different benchmarks. \textbf{AgentCPM-GUI}~\citep{zhang2025agentcpm} achieves leading performance on Chinese Android applications through cross-lingual training, progressive fine-tuning, and efficient on-device execution. \textbf{UI-TARS}~\citep{qin2025ui} adopts a purely vision-driven, end-to-end architecture that surpasses even closed-source models like GPT-4o, supported by large-scale action datasets and self-iterative training. \textbf{GUI-Owl}~\citep{ye2025mobile}, which serves as the backbone model for the Mobile-Agent-v3~\citep{ye2025mobile} framework, establishes new state-of-the-art results across both desktop and mobile environments by integrating scalable environment infrastructure, self-evolving trajectory production, and reinforcement learning. All models are deployed following their official guide, detailed descriptions are provided in the Appendix~\ref{app:models}.

\begin{figure}[h]
    \centering
    \begin{subfigure}{0.48\linewidth}
        \centering
        \includegraphics[width=\linewidth]{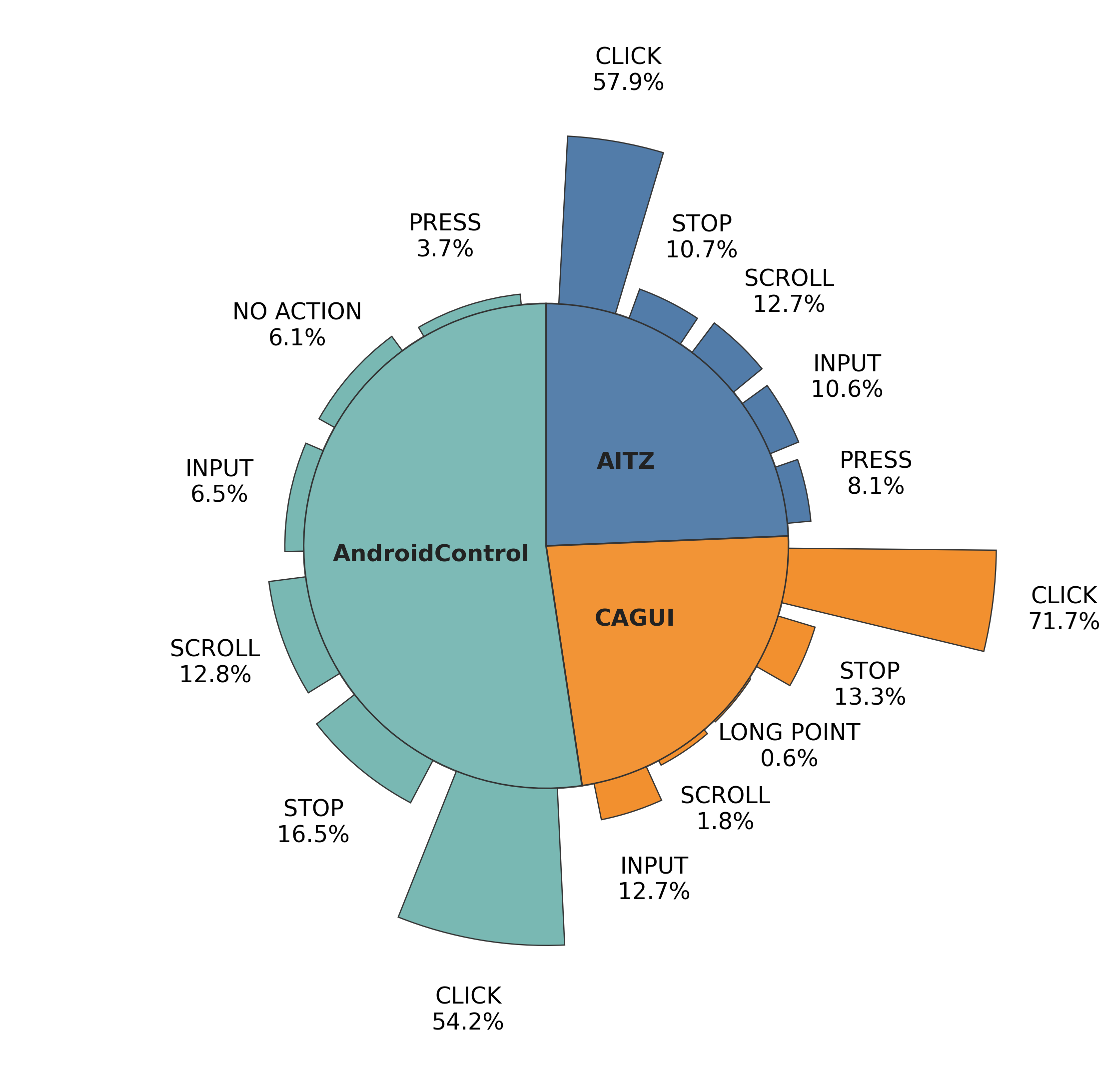}
    \end{subfigure}
    \hfill
    \begin{subfigure}{0.48\linewidth}
        \centering
        \includegraphics[width=\linewidth]{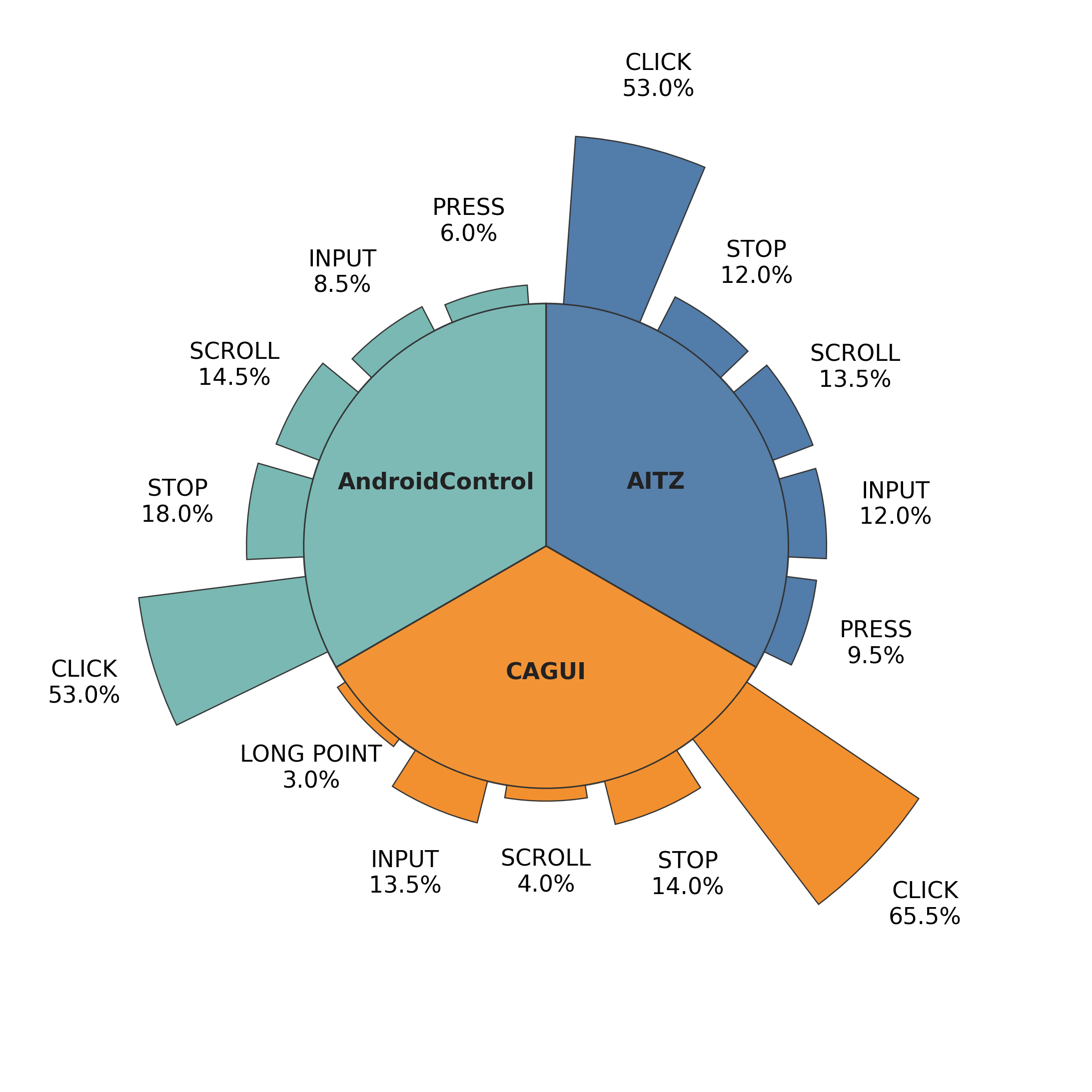}
    \end{subfigure}
    \vspace{-0.5em}
    \caption{Action distributions of the original datasets and the stratified sampled subset. 
    Overall, our sampling procedure preserves the overall distribution of actions while also ensuring that representative minority cases are adequately covered. 
    Left shows the full dataset distributions, while right illustrates the 1,800 sampled instances used for human annotation and agreement analysis.}
    \label{fig:dataset-comparison}
\end{figure}

\paragraph{Datasets}
We evaluate the above models on three benchmarks: AITZ~\citep{zhang2024android}, CAGUI~\citep{zhang2025agentcpm}, and AndroidControl~\citep{li2024effects}, where we adopt the high-level instruction split of AndroidControl, as it is more suitable for evaluating CoT reasoning quality. For each dataset, we collect the model-generated CoTs and predicted actions as evaluation data. To assess the reliability of our automated evaluation (RQ1), we apply stratified sampling across models and datasets to construct a subset of 1,800 instances for human annotation and agreement analysis. Our sampling procedure preserves the overall distribution of actions while also ensuring that representative minority cases are adequately covered. Figure~\ref{fig:dataset-comparison} illustrates the action distribution of the original datasets and the manually sampled subset, showing that the sampling preserves the overall distributional characteristics. Detailed descriptions of the datasets are provided in the Appendix~\ref{app:datasets}.

\vspace{-0.5em}

\paragraph{GTA Evaluator}
For the implementation of $f(c_n)$, we adopt AgentCPM-GUI-8B~\citep{zhang2025agentcpm} as the instruction-following VLM. 
We choose AgentCPM-GUI-8B~\citep{zhang2025agentcpm} because it is an open-source multimodal agent trained on large-scale GUI trajectories and paired reasoning data, 
which makes it particularly suitable for grounding free-form CoT into executable actions. 
During evaluation, we apply deterministic decoding to obtain $f(c_n)$ for each reasoning trace and compare the predicted actions against the ground-truth labels under the same strict matching rule as EM. 
This setup allows us to fairly assess the consistency between model reasoning and execution across different benchmarks.

\subsection{Evaluator Reliability (RQ1)}
\begin{figure}
    \centering
    \includegraphics[width=\linewidth]{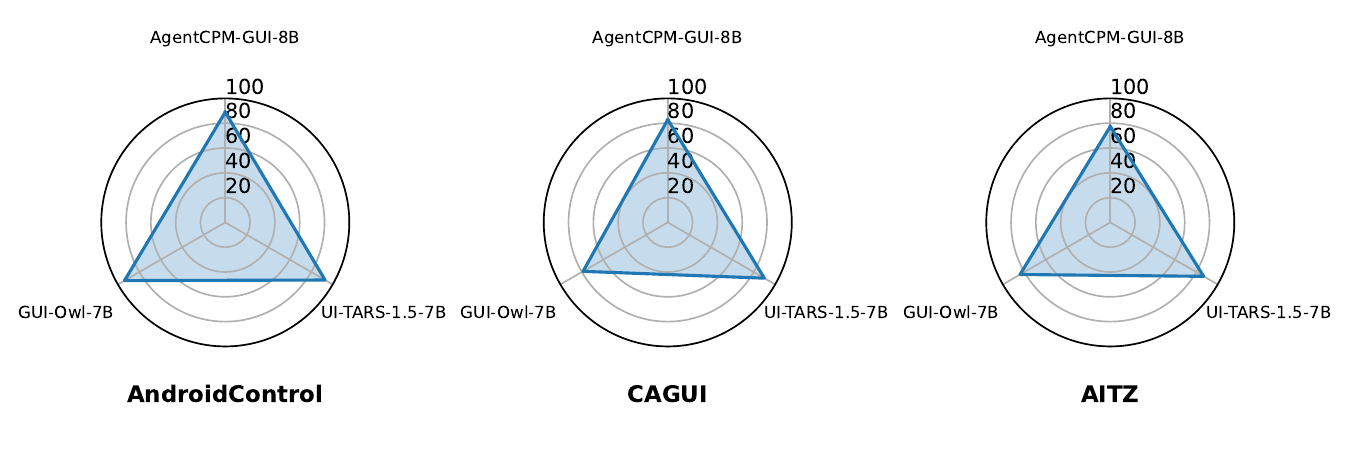}
    \vspace{-1cm}
    \caption{
    Radar plots show the \textbf{GTA evaluator accuracy} across three models and datasets. 
    Overall, the evaluator achieves consistently high accuracy, with similar performance across models. 
    Accuracy peaks on AndroidControl, while results on CAGUI and AITZ are slightly lower.
    }
    \label{fig:radar}
    \vspace{-0.4cm}
\end{figure}
\vspace{-0.4cm}
Since the evaluator is designed to automatically map free-form CoTs into their implied actions and assess their correctness, it is crucial to examine whether such automatic judgments align with human annotations. 
Establishing the reliability of the evaluator ensures that subsequent large-scale evaluations and comparative studies rest on a solid methodological foundation. 
To assess reliability, we first construct a human-annotated reference set by sampling CoTs and labeling their GTA outcomes following the procedure described below. 

Annotators are presented with the interface screenshot (with the ground-truth click-type action highlighted), the natural language instruction, the model-generated CoT, and the textual description of the ground-truth action. 
Given these inputs, annotators directly determine the $\mathrm{GTA}$ label according to the following criteria: 
(1) $\mathrm{GTA}=1$ if the CoT correctly implies the ground-truth action; 
(2) $\mathrm{GTA}=0$ if the CoT implies a different or incorrect action; 
(3) $\mathrm{GTA}=\mathrm{NA}$ if the case is undecidable, typically due to erroneous ground-truth annotations or severely incomplete CoTs. 

Each sample is independently annotated by two human experts. 
We retain only those instances where both annotators reach consensus on $\mathrm{GTA}\in\{0,1\}$. 
Samples for which at least one annotator assigns $\mathrm{NA}$, or where the two annotators disagree, are discarded. 
This procedure ensures that the resulting human-annotated subset maintains high reliability and provides a solid reference for evaluating the automatic $\mathrm{GTA}$ evaluator. 

We then apply the GTA Evaluator to the same samples and compute its prediction accuracy with respect to human consensus labels. This approach allows us to directly measure the degree to which the automatic evaluator replicates human judgment. 

Radar plots in Figure~\ref{fig:radar} illustrate the accuracy of the GTA Evaluator across three models and three datasets. 
Overall, the evaluator achieves consistently high accuracy, with only minor variation across different models. 
Among the datasets, accuracy peaks on AndroidControl~\citep{li2024effects}, while performance on CAGUI~\citep{zhang2025agentcpm} and AITZ~\citep{zhang2024android} is slightly lower. 
These results suggest that the evaluator provides a stable and trustworthy proxy for human assessment. 

The analysis indicates that the GTA Evaluator generalizes well across both models and datasets, reducing the need for costly and time-consuming human annotations. 
While minor dataset-specific differences remain, the overall robustness of the evaluator highlights its potential as a scalable evaluation protocol for future research. 
Beyond the current study, these findings point toward broader applications in automated assessment of reasoning-action alignment, where reliable and reproducible metrics are essential. 

\subsection{Model Performance Assessment (RQ2)}

\begin{figure}
    \centering
    \includegraphics[width=1\linewidth]{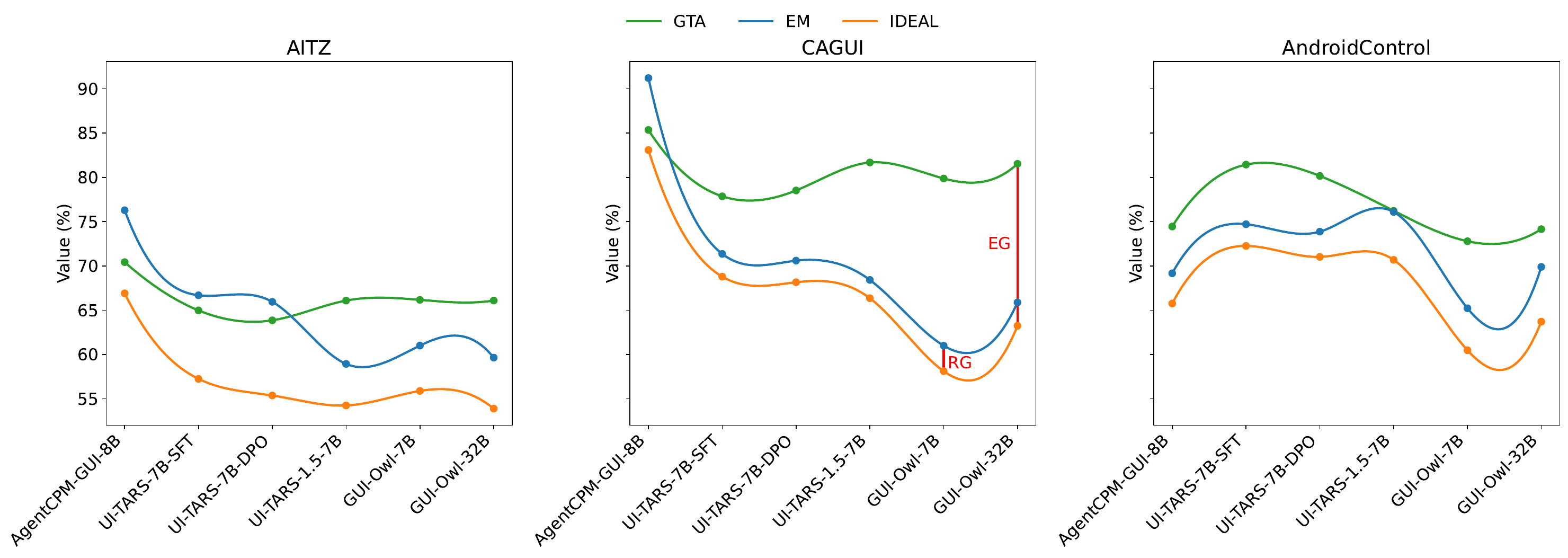}
    \caption{Spline plots of $\mathrm{GTA}$, $\mathrm{EM}$, and $\mathrm{IDEAL}$. 
    $\mathrm{EM}$ measures execution accuracy, $\mathrm{GTA}$ reflects reasoning accuracy, and $\mathrm{IDEAL}$ means ideal reasoning and execution. 
    By construction, $\mathrm{GTA}-\mathrm{IDEAL}=\mathrm{EG}$ and $\mathrm{EM}-\mathrm{IDEAL}=\mathrm{RG}$. 
    When $\mathrm{GTA}$ lies above $\mathrm{EM}$, it indicates $\mathrm{EG}>\mathrm{RG}$, revealing that the main bottleneck lies in translating correct reasoning into executable actions.}
    \label{fig:spline}
    \vspace{-0.4cm}
\end{figure}
\vspace{-0.5em}

Execution Match (EM) and Ground Truth Alignment (GTA) capture two complementary perspectives of model performance: 
EM measures strict execution accuracy, while GTA reflects whether the reasoning consistently implies the correct action. 
Evaluating them jointly is crucial, since high EM does not necessarily imply high GTA, and vice versa. 
We therefore examine EM and GTA across three benchmarks to assess both execution reliability and reasoning alignment. 

We report EM and GTA scores for six representative models on AITZ~\citep{zhang2024android}, CAGUI~\citep{zhang2025agentcpm}, and AndroidControl~\citep{li2024effects}, where we adopt the high-level instruction split of AndroidControl, 
as it is more suitable for evaluating CoT reasoning quality. 
Results are visualized with grouped bar charts, where each dataset provides paired comparisons of EM and GTA. 
This setup allows us to jointly interpret execution accuracy and reasoning accuracy, and to further inspect their divergence. 

Overall, EM and GTA exhibit similar patterns across benchmarks, but important divergences emerge. 
From Figure~\ref{fig:spline}, we have the following key findings.

\paragraph{Bottleneck lies in EG.} Across 18 model-dataset combinations, GTA surpasses EM in 14 cases, indicating that $\mathrm{EG}>\mathrm{RG}$ in the majority of scenarios.
This suggests that most models can \emph{reason correctly}, at least at the level of generating valid chains of thought, but encounter difficulty when mapping these intermediate reasoning steps into the precise executable actions required by the task.
In other words, the primary bottleneck lies not in reasoning correctness itself, but in the reasoning-to-execution transition, where small misalignments or ambiguities in action formulation often lead to task failure.
Such results highlight the importance of improving models’ ability to faithfully ground abstract reasoning in concrete actions, an ability that becomes especially critical in domains requiring fine-grained decision-making.

\paragraph{Causal CoT Models often exhibit $\mathrm{RG}>\mathrm{EG}$.} We also observe counterexamples where EM exceeds GTA (e.g., AITZ results of AgentCPM and UI-TARS).  
In these cases, models rely heavily on action shortcuts learned during supervised fine-tuning, 
while ignoring or even contradicting their CoT reasoning. 
This phenomenon is particularly pronounced on AITZ, whose long and sometimes inconsistent CoTs amplify the problem: 
if the model’s reasoning ability is almost entirely inherited from SFT, it tends to overfit to actions and disregard CoT consistency.

\paragraph{OOD data highlight different bottlenecks.}
On the out-of-distribution benchmark CAGUI, untrained models consistently show high GTA but very large EG, 
suggesting that the main challenge lies in \emph{grounding reasoning into unfamiliar screens}.  
Interestingly, the AgentCPM-GUI model, although achieving the best EM and GTA on CAGUI, 
still exhibits a large RG, indicating that its training on Chinese GUI data has encouraged reliance on action shortcuts.  
This observation highlights that stronger training can improve execution performance but may simultaneously exacerbate reasoning-faithfulness issues.

Together, these insights show that combination of EM and GTA not only benchmark accuracy, 
but also reveal different failure modes: translation bottlenecks (high EG), shortcut reliance (high RG), 
and grounding challenges (large EG on OOD data).  
Such analysis highlights the necessity of joint metrics in diagnosing reasoning-execution alignment.

Representative qualitative examples for all four quadrants are provided in Appendix~\ref{appendix:case}.

\subsection{Parameter Scaling Effects (RQ3)}
\begin{figure}[ht]
    \centering
    \begin{subfigure}{0.48\linewidth}
        \centering
        \includegraphics[width=\linewidth]{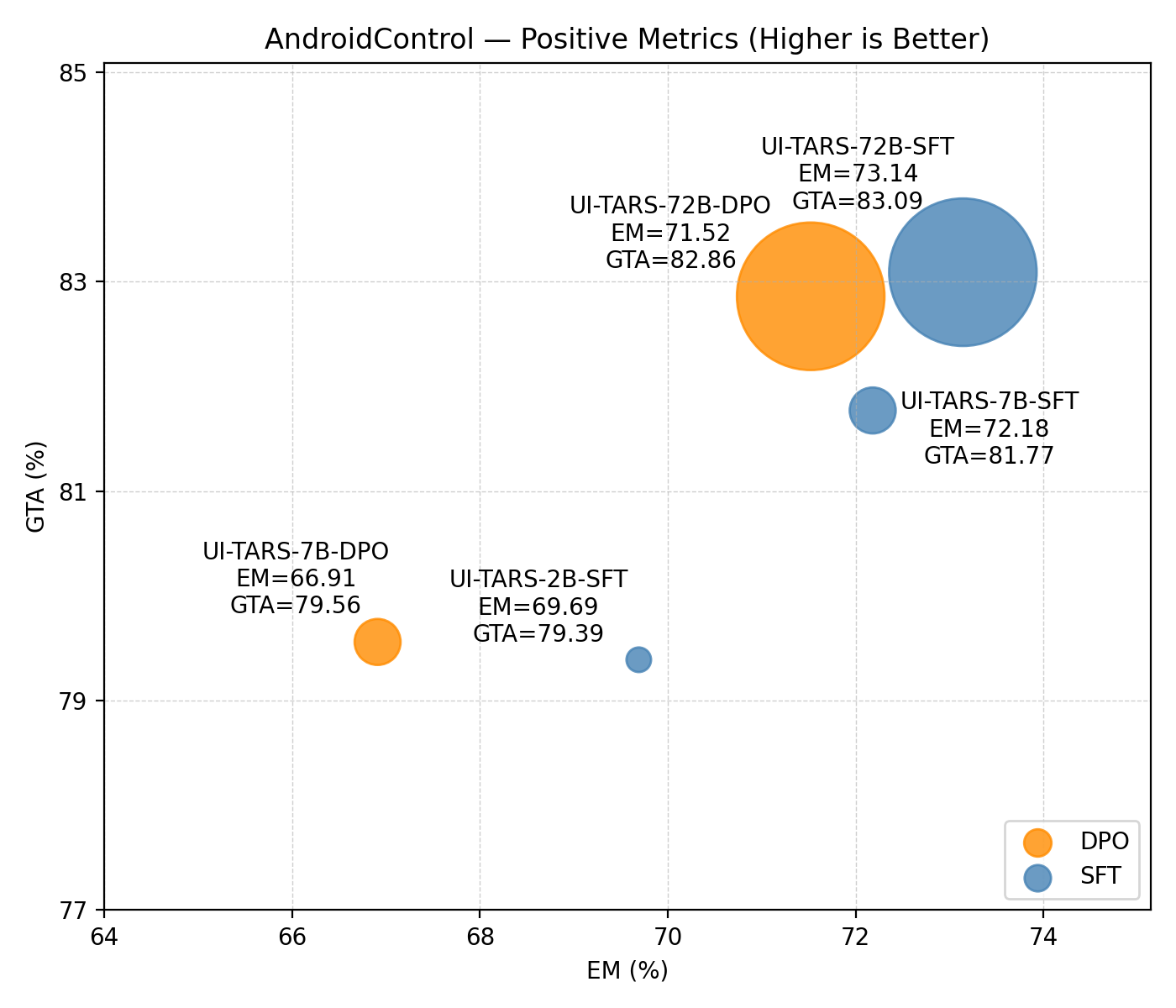}
        \caption{Positive Metrics}
    \end{subfigure}
    \hfill
    \begin{subfigure}{0.48\linewidth}
        \centering
        \includegraphics[width=\linewidth]{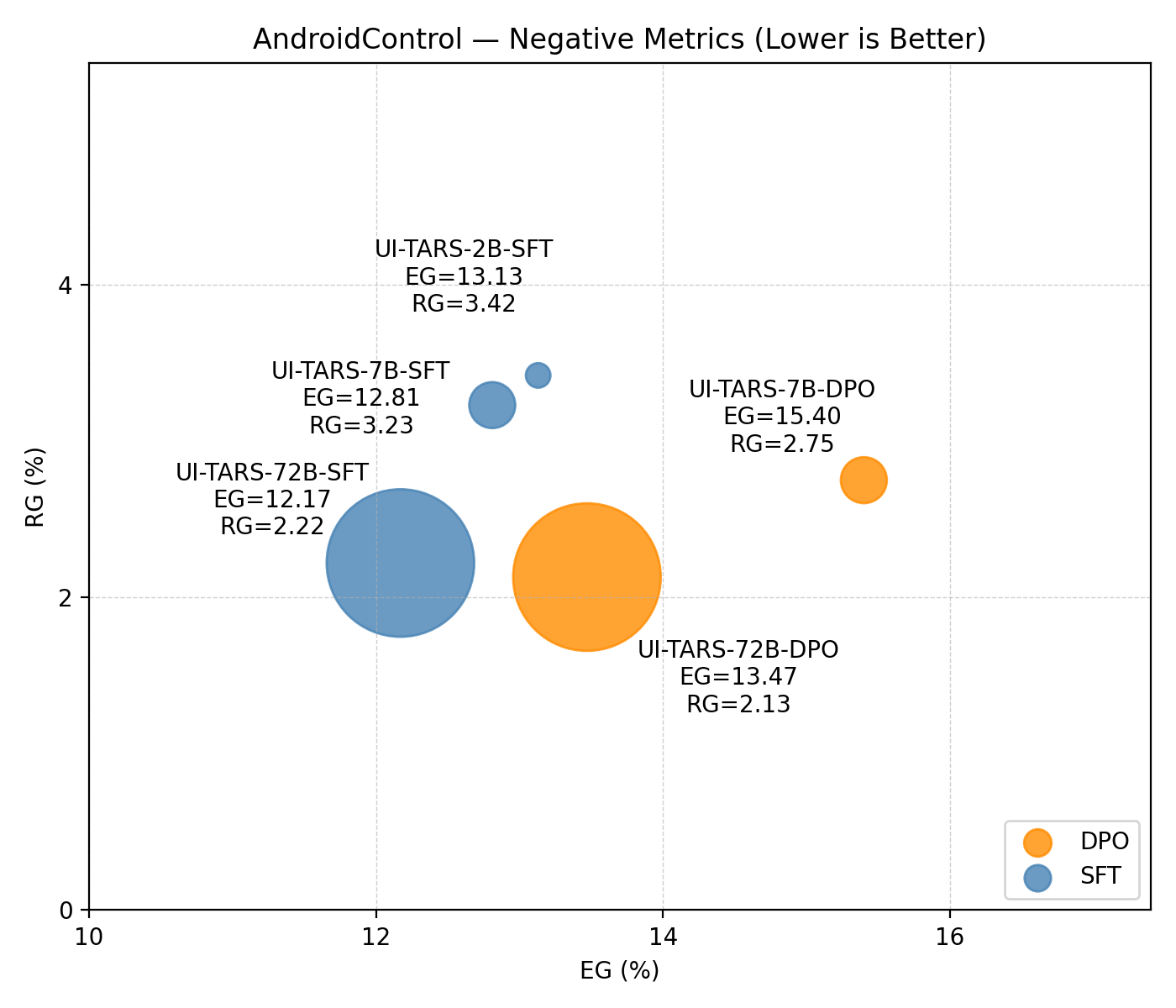}
        \caption{Negative Metrics}
    \end{subfigure}
    \vspace{-0.5em}
    \caption{Effect of parameter scaling on reasoning--execution gaps in AndroidControl. 
(a) Positive metrics: EM and GTA, where higher is better. 
(b) Negative metrics: EG and RG, where lower is better. 
Orange points denote DPO models and blue points denote SFT models, with point size proportional to parameter scale. 
Scaling consistently improves EM and GTA while reducing EG and RG, though even the largest (72B) model still exhibits execution gaps above 10\%.}
    \label{fig:scaling}
\end{figure}
\vspace{-0.5em}
A central question in multimodal agent research is whether increasing model scale 
can effectively alleviate the reasoning-execution gap. 
While larger parameter counts typically enhance general reasoning abilities of large language models, 
it remains unclear whether such improvements translate to consistent gains in GUI interaction tasks, 
particularly in reducing execution errors.

We systematically evaluate UI-Tars~\citep{qin2025ui} models of varying sizes (2B, 7B, and 72B) under two training paradigms 
(\textit{SFT} and \textit{DPO}) on the AndroidControl benchmark. 
We analyze their performance using four metrics: 
Execution Match (EM) and Ground-Truth Alignment (GTA) as positive indicators, 
and Execution Gap (EG) and Reasoning Gap (RG) as negative indicators. 
Visualization is provided in Figure~\ref{fig:scaling}, 
where point size encodes parameter scale.

As shown in Figure~\ref{fig:scaling}, scaling consistently improves both EM and GTA, 
with larger models achieving better alignment between predicted and ground-truth actions. 
At the same time, EG and RG decrease monotonically with parameter growth, 
indicating that scaling narrows the reasoning--execution gap. 
Notably, however, the largest 72B models still exhibit residual execution gaps ($>$10\%), 
suggesting that scaling alone cannot fully eliminate misalignment. 
We also observe comparable trends across training paradigms, 
though DPO models achieve slightly stronger alignment than SFT models at the same scale.

These findings confirm that parameter scaling enhances both reasoning quality 
and execution accuracy in multimodal GUI agents, 
yet highlight diminishing returns in closing the reasoning-execution-gaps. 
This suggests that future work must move beyond pure scaling 
to incorporate training strategies or architectural designs 
that directly target reasoning-action consistency.

\section{Conclusion}
In this work, we analyzed the reasoning-execution gaps in VLM-powered mobile-use agents. We introduced a framework to disentangle reasoning from execution in VLM-powered mobile-use agents. In contrast to prior work emphasizing action accuracy, our Ground-Truth Alignment (GTA) metric exposes reasoning–execution gaps through a four-quadrant diagnostic that highlights overlooked failure modes. Evaluations on three mobile benchmarks show that these gaps are common, with execution gaps dominating even in strong models. While scaling reduces misalignment, execution gaps remain, suggesting that progress cannot rely on size alone. 
By shifting evaluation from execution accuracy to reasoning–execution gaps, this work offers a step toward more transparent and trustworthy assessments of mobile-use agents.
\clearpage
\newpage

\section*{Ethics Statement} 
This work studies reasoning–execution gaps in VLM-powered mobile-use agents. The models evaluated in this work were obtained from publicly available official repositories and used strictly under the licenses and usage terms specified by the original authors. All datasets employed in our experiments are publicly available and cited appropriately. For the small portion of data requiring human annotation, annotators were instructed to avoid recording any personal or sensitive information, and only samples with full agreement between annotators were retained to ensure reliability. Proper citations are included throughout the paper to ensure attribution and transparency.

\section*{Reproducibility}
To facilitate reproducibility, we provide detailed descriptions of model deployment and evaluation 
procedures in the Appendix~\ref{app:models} and Appendix~\ref{app:evaluation}. During the submission phase, we have uploaded supplementary code 
with all author-identifying information anonymized. The supplementary materials include 
instructions for setting up environments, running the GTA evaluator, and replicating our 
experiments across the three benchmarks. Upon acceptance, we will publicly release the full 
codebase and scripts to ensure transparency and enable independent verification of our findings.



\bibliography{iclr2026_conference}

\begin{thebibliography}{60}
\providecommand{\natexlab}[1]{#1}
\providecommand{\url}[1]{\texttt{#1}}
\expandafter\ifx\csname urlstyle\endcsname\relax
  \providecommand{\doi}[1]{doi: #1}\else
  \providecommand{\doi}{doi: \begingroup \urlstyle{rm}\Url}\fi

\bibitem[Agashe et~al.(2025)Agashe, Wong, Tu, Yang, Li, and Wang]{agashe2025agent}
Saaket Agashe, Kyle Wong, Vincent Tu, Jiachen Yang, Ang Li, and Xin~Eric Wang.
\newblock Agent s2: A compositional generalist-specialist framework for computer use agents.
\newblock \emph{arXiv preprint arXiv:2504.00906}, 2025.

\bibitem[Ai et~al.(2025)Ai, Bu, Cao, Wang, Gu, Xing, Zhu, Jiang, Zheng, Song, et~al.]{ai2025inquiremobile}
Qihang Ai, Pi~Bu, Yue Cao, Yingyao Wang, Jihao Gu, Jingxuan Xing, Zekun Zhu, Wei Jiang, Zhicheng Zheng, Jun Song, et~al.
\newblock Inquiremobile: Teaching vlm-based mobile agent to request human assistance via reinforcement fine-tuning.
\newblock \emph{arXiv preprint arXiv:2508.19679}, 2025.

\bibitem[Bai et~al.(2025)Bai, Chen, Liu, Wang, Ge, Song, Dang, Wang, Wang, Tang, et~al.]{bai2025qwen2}
Shuai Bai, Keqin Chen, Xuejing Liu, Jialin Wang, Wenbin Ge, Sibo Song, Kai Dang, Peng Wang, Shijie Wang, Jun Tang, et~al.
\newblock Qwen2. 5-vl technical report.
\newblock \emph{arXiv preprint arXiv:2502.13923}, 2025.
\newblock \doi{10.48550/arXiv.2502.13923}.

\bibitem[Barez et~al.(2025)Barez, Wu, Arcuschin, Lan, Wang, Siegel, Collignon, Neo, Lee, Paren, et~al.]{barez2025chain}
Fazl Barez, Tung-Yu Wu, Iv{\'a}n Arcuschin, Michael Lan, Vincent Wang, Noah Siegel, Nicolas Collignon, Clement Neo, Isabelle Lee, Alasdair Paren, et~al.
\newblock Chain-of-thought is not explainability.
\newblock \emph{Preprint, alphaXiv}, pp.\ ~v2, 2025.

\bibitem[Chen et~al.(2024)Chen, Yuen, Xie, Yang, Chen, Wu, Yixing, Zhou, Liu, Wang, et~al.]{chen2024spa}
Jingxuan Chen, Derek Yuen, Bin Xie, Yuhao Yang, Gongwei Chen, Zhihao Wu, Li~Yixing, Xurui Zhou, Weiwen Liu, Shuai Wang, et~al.
\newblock Spa-bench: A comprehensive benchmark for smartphone agent evaluation.
\newblock In \emph{NeurIPS 2024 Workshop on Open-World Agents}, 2024.

\bibitem[Chen et~al.(2025{\natexlab{a}})Chen, Benton, Radhakrishnan, Uesato, Denison, Schulman, Somani, Hase, Wagner, Roger, et~al.]{chen2505reasoning}
Yanda Chen, Joe Benton, Ansh Radhakrishnan, Jonathan Uesato, Carson Denison, John Schulman, Arushi Somani, Peter Hase, Misha Wagner, Fabien Roger, et~al.
\newblock Reasoning models don’t always say what they think.
\newblock \emph{URL: https://arxiv. org/abs/2505.05410}, 2025{\natexlab{a}}.

\bibitem[Chen et~al.(2025{\natexlab{b}})Chen, Hu, Yin, Li, and Zhang]{chen2025aeia}
Yurun Chen, Xueyu Hu, Keting Yin, Juncheng Li, and Shengyu Zhang.
\newblock Aeia-mn: Evaluating the robustness of multimodal llm-powered mobile agents against active environmental injection attacks.
\newblock \emph{arXiv e-prints}, pp.\  arXiv--2502, 2025{\natexlab{b}}.

\bibitem[Cheng et~al.(2024)Cheng, Sun, Chu, Xu, Li, Zhang, and Wu]{cheng2024seeclick}
Kanzhi Cheng, Qiushi Sun, Yougang Chu, Fangzhi Xu, Yantao Li, Jianbing Zhang, and Zhiyong Wu.
\newblock Seeclick: Harnessing gui grounding for advanced visual gui agents.
\newblock \emph{arXiv preprint arXiv:2401.10935}, 2024.

\bibitem[Cheng et~al.(2025{\natexlab{a}})Cheng, Wu, Wu, Ju, Zhang, Zhang, and Liu]{cheng-etal-2025-os}
Pengzhou Cheng, Zheng Wu, Zongru Wu, Tianjie Ju, Aston Zhang, Zhuosheng Zhang, and Gongshen Liu.
\newblock {OS}-kairos: Adaptive interaction for {MLLM}-powered {GUI} agents.
\newblock In Wanxiang Che, Joyce Nabende, Ekaterina Shutova, and Mohammad~Taher Pilehvar (eds.), \emph{Findings of the Association for Computational Linguistics: ACL 2025}, pp.\  6701--6725, Vienna, Austria, July 2025{\natexlab{a}}. Association for Computational Linguistics.
\newblock ISBN 979-8-89176-256-5.
\newblock \doi{10.18653/v1/2025.findings-acl.348}.
\newblock URL \url{https://aclanthology.org/2025.findings-acl.348/}.

\bibitem[Cheng et~al.(2025{\natexlab{b}})Cheng, Chen, Xu, Wang, Wang, Fei, Wang, Wang, Chen, Che, et~al.]{cheng2025visual}
Zihui Cheng, Qiguang Chen, Xiao Xu, Jiaqi Wang, Weiyun Wang, Hao Fei, Yidong Wang, Alex~Jinpeng Wang, Zhi Chen, Wanxiang Che, et~al.
\newblock Visual thoughts: A unified perspective of understanding multimodal chain-of-thought.
\newblock \emph{arXiv preprint arXiv:2505.15510}, 2025{\natexlab{b}}.

\bibitem[Cheng et~al.(2025{\natexlab{c}})Cheng, Chen, Zhang, Fei, Feng, Che, Li, and Qin]{cheng2025comt}
Zihui Cheng, Qiguang Chen, Jin Zhang, Hao Fei, Xiaocheng Feng, Wanxiang Che, Min Li, and Libo Qin.
\newblock Comt: A novel benchmark for chain of multi-modal thought on large vision-language models.
\newblock In \emph{Proceedings of the AAAI Conference on Artificial Intelligence}, volume~39, pp.\  23678--23686, 2025{\natexlab{c}}.

\bibitem[Cheng et~al.(2025{\natexlab{d}})Cheng, Huang, Pan, Hou, and Zhan]{cheng2025navi}
Ziming Cheng, Zhiyuan Huang, Junting Pan, Zhaohui Hou, and Mingjie Zhan.
\newblock Navi-plus: Managing ambiguous gui navigation tasks with follow-up.
\newblock \emph{arXiv preprint arXiv:2503.24180}, 2025{\natexlab{d}}.

\bibitem[Comanici et~al.(2025)Comanici, Bieber, Schaekermann, Pasupat, Sachdeva, Dhillon, Blistein, Ram, Zhang, Rosen, et~al.]{comanici2025gemini}
Gheorghe Comanici, Eric Bieber, Mike Schaekermann, Ice Pasupat, Noveen Sachdeva, Inderjit Dhillon, Marcel Blistein, Ori Ram, Dan Zhang, Evan Rosen, et~al.
\newblock Gemini 2.5: Pushing the frontier with advanced reasoning, multimodality, long context, and next generation agentic capabilities.
\newblock \emph{arXiv preprint arXiv:2507.06261}, 2025.
\newblock URL \url{https://arxiv.org/abs/2507.06261}.

\bibitem[Gou et~al.(2024)Gou, Wang, Zheng, Xie, Chang, Shu, Sun, and Su]{gou2024navigating}
Boyu Gou, Ruohan Wang, Boyuan Zheng, Yanan Xie, Cheng Chang, Yiheng Shu, Huan Sun, and Yu~Su.
\newblock Navigating the digital world as humans do: Universal visual grounding for gui agents.
\newblock \emph{arXiv preprint arXiv:2410.05243}, 2024.

\bibitem[Gu et~al.(2025{\natexlab{a}})Gu, Ai, Wang, Bu, Xing, Zhu, Jiang, Wang, Zhao, Zhang, et~al.]{gu2025mobile}
Jihao Gu, Qihang Ai, Yingyao Wang, Pi~Bu, Jingxuan Xing, Zekun Zhu, Wei Jiang, Ziming Wang, Yingxiu Zhao, Ming-Liang Zhang, et~al.
\newblock Mobile-r1: Towards interactive reinforcement learning for vlm-based mobile agent via task-level rewards.
\newblock \emph{arXiv preprint arXiv:2506.20332}, 2025{\natexlab{a}}.

\bibitem[Gu et~al.(2025{\natexlab{b}})Gu, Zeng, Xu, Zhou, Shen, Liu, Zhou, Meng, Xia, Chen, et~al.]{gu2025ui}
Zhangxuan Gu, Zhengwen Zeng, Zhenyu Xu, Xingran Zhou, Shuheng Shen, Yunfei Liu, Beitong Zhou, Changhua Meng, Tianyu Xia, Weizhi Chen, et~al.
\newblock Ui-venus technical report: Building high-performance ui agents with rft.
\newblock \emph{arXiv preprint arXiv:2508.10833}, 2025{\natexlab{b}}.

\bibitem[Guo et~al.(2025)Guo, Miao, Wu, Cheng, Zhou, and Zhang]{guo2025atomic}
Yuan Guo, Tingjia Miao, Zheng Wu, Pengzhou Cheng, Ming Zhou, and Zhuosheng Zhang.
\newblock Atomic-to-compositional generalization for mobile agents with a new benchmark and scheduling system.
\newblock \emph{arXiv preprint arXiv:2506.08972}, 2025.

\bibitem[Hao et~al.(2025)Hao, Wang, and Zhou]{hao2025uncertainty}
Chao Hao, Shuai Wang, and Kaiwen Zhou.
\newblock Uncertainty-aware gui agent: Adaptive perception through component recommendation and human-in-the-loop refinement.
\newblock \emph{arXiv preprint arXiv:2508.04025}, 2025.

\bibitem[Hong et~al.(2024)Hong, Wang, Lv, Xu, Yu, Ji, Wang, Wang, Dong, Ding, et~al.]{hong2024cogagent}
Wenyi Hong, Weihan Wang, Qingsong Lv, Jiazheng Xu, Wenmeng Yu, Junhui Ji, Yan Wang, Zihan Wang, Yuxiao Dong, Ming Ding, et~al.
\newblock Cogagent: A visual language model for gui agents.
\newblock In \emph{Proceedings of the IEEE/CVF Conference on Computer Vision and Pattern Recognition}, pp.\  14281--14290, 2024.

\bibitem[Hu et~al.(2025)Hu, Xiong, Yi, Wei, Xiao, Chen, Ye, Tao, Zhou, Zhao, et~al.]{hu2025agents}
Xueyu Hu, Tao Xiong, Biao Yi, Zishu Wei, Ruixuan Xiao, Yurun Chen, Jiasheng Ye, Meiling Tao, Xiangxin Zhou, Ziyu Zhao, et~al.
\newblock Os agents: A survey on mllm-based agents for general computing devices use.
\newblock \emph{arXiv preprint arXiv:2508.04482}, 2025.

\bibitem[Li et~al.(2025)Li, Qu, Zhou, Wang, Wen, Du, Lou, Peng, and Zhang]{li2025mobileuse}
Ning Li, Xiangmou Qu, Jiamu Zhou, Jun Wang, Muning Wen, Kounianhua Du, Xingyu Lou, Qiuying Peng, and Weinan Zhang.
\newblock Mobileuse: A gui agent with hierarchical reflection for autonomous mobile operation.
\newblock \emph{arXiv preprint arXiv:2507.16853}, 2025.

\bibitem[Li et~al.(2024)Li, Bishop, Li, Rawles, Campbell-Ajala, Tyamagundlu, and Riva]{li2024effects}
Wei Li, William~E Bishop, Alice Li, Christopher Rawles, Folawiyo Campbell-Ajala, Divya Tyamagundlu, and Oriana Riva.
\newblock On the effects of data scale on ui control agents.
\newblock \emph{Advances in Neural Information Processing Systems}, 37:\penalty0 92130--92154, 2024.

\bibitem[Liu et~al.(2025{\natexlab{a}})Liu, Zhao, Liu, Guo, Xiao, Lin, Chai, Han, Ren, Wang, et~al.]{liu2025llm}
Guangyi Liu, Pengxiang Zhao, Liang Liu, Yaxuan Guo, Han Xiao, Weifeng Lin, Yuxiang Chai, Yue Han, Shuai Ren, Hao Wang, et~al.
\newblock Llm-powered gui agents in phone automation: Surveying progress and prospects.
\newblock \emph{arXiv preprint arXiv:2504.19838}, 2025{\natexlab{a}}.

\bibitem[Liu et~al.(2025{\natexlab{b}})Liu, Li, Xie, Hu, Han, Zhang, Yang, and Wu]{liu2025infigui}
Yuhang Liu, Pengxiang Li, Congkai Xie, Xavier Hu, Xiaotian Han, Shengyu Zhang, Hongxia Yang, and Fei Wu.
\newblock Infigui-r1: Advancing multimodal gui agents from reactive actors to deliberative reasoners.
\newblock \emph{arXiv preprint arXiv:2504.14239}, 2025{\natexlab{b}}.

\bibitem[Lu et~al.(2025)Lu, Chai, Guo, Yin, Liu, Wang, Xiao, Ren, Xiong, and Li]{lu2025ui}
Zhengxi Lu, Yuxiang Chai, Yaxuan Guo, Xi~Yin, Liang Liu, Hao Wang, Han Xiao, Shuai Ren, Guanjing Xiong, and Hongsheng Li.
\newblock Ui-r1: Enhancing efficient action prediction of gui agents by reinforcement learning.
\newblock \emph{arXiv preprint arXiv:2503.21620}, 2025.

\bibitem[Luo et~al.(2025)Luo, Wang, He, and Xia]{luo2025gui}
Run Luo, Lu~Wang, Wanwei He, and Xiaobo Xia.
\newblock Gui-r1: A generalist r1-style vision-language action model for gui agents.
\newblock \emph{arXiv preprint arXiv:2504.10458}, 2025.

\bibitem[Ma et~al.(2025)Ma, Wang, Yao, Yuan, Zhang, Zhang, and Zhao]{ma2025caution}
Xinbei Ma, Yiting Wang, Yao Yao, Tongxin Yuan, Aston Zhang, Zhuosheng Zhang, and Hai Zhao.
\newblock Caution for the environment: Multimodal llm agents are susceptible to environmental distractions.
\newblock In \emph{Proceedings of the 63rd Annual Meeting of the Association for Computational Linguistics (Volume 1: Long Papers)}, pp.\  22324--22339, 2025.

\bibitem[Matton et~al.(2025)Matton, Ness, Guttag, and K{\i}c{\i}man]{matton2025walk}
Katie Matton, Robert~Osazuwa Ness, John Guttag, and Emre K{\i}c{\i}man.
\newblock Walk the talk? measuring the faithfulness of large language model explanations.
\newblock \emph{arXiv preprint arXiv:2504.14150}, 2025.

\bibitem[Nguyen et~al.(2024)Nguyen, Chen, Wang, Wu, Park, Hu, Lyu, Wu, Aponte, Xia, et~al.]{nguyen2024gui}
Dang Nguyen, Jian Chen, Yu~Wang, Gang Wu, Namyong Park, Zhengmian Hu, Hanjia Lyu, Junda Wu, Ryan Aponte, Yu~Xia, et~al.
\newblock Gui agents: A survey.
\newblock \emph{arXiv preprint arXiv:2412.13501}, 2024.

\bibitem[OpenAI(2024)]{openai2024gpt4o}
OpenAI.
\newblock Gpt-4 system card.
\newblock 2024.
\newblock URL \url{https://cdn.openai.com/papers/gpt-4-system-card.pdf}.

\bibitem[OpenAI(2025)]{openai2025gpt5}
OpenAI.
\newblock Gpt-5 system card.
\newblock 2025.
\newblock URL \url{https://cdn.openai.com/gpt-5-system-card.pdf}.

\bibitem[Qin et~al.(2025)Qin, Ye, Fang, Wang, Liang, Tian, Zhang, Li, Li, Huang, et~al.]{qin2025ui}
Yujia Qin, Yining Ye, Junjie Fang, Haoming Wang, Shihao Liang, Shizuo Tian, Junda Zhang, Jiahao Li, Yunxin Li, Shijue Huang, et~al.
\newblock Ui-tars: Pioneering automated gui interaction with native agents.
\newblock \emph{arXiv preprint arXiv:2501.12326}, 2025.

\bibitem[Shi et~al.(2025)Shi, Yu, Yao, Chen, and Liu]{shi2025towards}
Yucheng Shi, Wenhao Yu, Wenlin Yao, Wenhu Chen, and Ninghao Liu.
\newblock Towards trustworthy gui agents: A survey.
\newblock \emph{arXiv preprint arXiv:2503.23434}, 2025.

\bibitem[Tang et~al.(2025)Tang, Dong, Huang, Xiang, Ruan, Wang, Li, Cao, Pang, Kong, et~al.]{tang2025magicgui}
Liujian Tang, Shaokang Dong, Yijia Huang, Minqi Xiang, Hongtao Ruan, Bin Wang, Shuo Li, Zhihui Cao, Hailiang Pang, Heng Kong, et~al.
\newblock Magicgui: A foundational mobile gui agent with scalable data pipeline and reinforcement fine-tuning.
\newblock \emph{arXiv preprint arXiv:2508.03700}, 2025.

\bibitem[Tao et~al.(2025)Tao, Wang, Cai, Yang, and Tang]{tao2025understanding}
Xingjian Tao, Yiwei Wang, Yujun Cai, Zhicheng Yang, and Jing Tang.
\newblock Understanding gui agent localization biases through logit sharpness.
\newblock \emph{arXiv preprint arXiv:2506.15425}, 2025.

\bibitem[Team et~al.(2024)Team, Georgiev, Lei, Burnell, Bai, Gulati, Tanzer, Vincent, Pan, Wang, et~al.]{team2024gemini}
Gemini Team, Petko Georgiev, Ving~Ian Lei, Ryan Burnell, Libin Bai, Anmol Gulati, Garrett Tanzer, Damien Vincent, Zhufeng Pan, Shibo Wang, et~al.
\newblock Gemini 1.5: Unlocking multimodal understanding across millions of tokens of context.
\newblock \emph{arXiv preprint arXiv:2403.05530}, 2024.
\newblock URL \url{https://doi.org/10.48550/arXiv.2403.05530}.

\bibitem[Wang et~al.(2024{\natexlab{a}})Wang, Xu, Jia, Zhang, Yan, Shen, Zhang, Huang, and Sang]{wang2024mobile}
Junyang Wang, Haiyang Xu, Haitao Jia, Xi~Zhang, Ming Yan, Weizhou Shen, Ji~Zhang, Fei Huang, and Jitao Sang.
\newblock Mobile-agent-v2: Mobile device operation assistant with effective navigation via multi-agent collaboration.
\newblock \emph{Advances in Neural Information Processing Systems}, 37:\penalty0 2686--2710, 2024{\natexlab{a}}.

\bibitem[Wang et~al.(2024{\natexlab{b}})Wang, Bai, Tan, Wang, Fan, Bai, Chen, Liu, Wang, Ge, et~al.]{wang2024qwen2}
Peng Wang, Shuai Bai, Sinan Tan, Shijie Wang, Zhihao Fan, Jinze Bai, Keqin Chen, Xuejing Liu, Jialin Wang, Wenbin Ge, et~al.
\newblock Qwen2-vl: Enhancing vision-language model's perception of the world at any resolution.
\newblock \emph{arXiv preprint arXiv:2409.12191}, 2024{\natexlab{b}}.
\newblock \doi{https://doi.org/10.48550/arXiv.2409.12191}.

\bibitem[Wang et~al.(2024{\natexlab{c}})Wang, Liu, Chen, Zhou, Gan, Zeng, Che, Yu, Hao, Shao, et~al.]{wang2024gui}
Shuai Wang, Weiwen Liu, Jingxuan Chen, Yuqi Zhou, Weinan Gan, Xingshan Zeng, Yuhan Che, Shuai Yu, Xinlong Hao, Kun Shao, et~al.
\newblock Gui agents with foundation models: A comprehensive survey.
\newblock \emph{arXiv preprint arXiv:2411.04890}, 2024{\natexlab{c}}.

\bibitem[Wang et~al.(2025)Wang, Wu, Xie, Ding, Yang, Li, Liu, Li, Dong, Chen, et~al.]{wang2025mmbench}
Xuehui Wang, Zhenyu Wu, JingJing Xie, Zichen Ding, Bowen Yang, Zehao Li, Zhaoyang Liu, Qingyun Li, Xuan Dong, Zhe Chen, et~al.
\newblock Mmbench-gui: Hierarchical multi-platform evaluation framework for gui agents.
\newblock \emph{arXiv preprint arXiv:2507.19478}, 2025.

\bibitem[Wang et~al.(2024{\natexlab{d}})Wang, Mao, Fried, and Neubig]{wang2024agent}
Zora~Zhiruo Wang, Jiayuan Mao, Daniel Fried, and Graham Neubig.
\newblock Agent workflow memory.
\newblock \emph{arXiv preprint arXiv:2409.07429}, 2024{\natexlab{d}}.

\bibitem[Wu et~al.(2025)Wu, Huang, Lou, Qu, Cheng, Wu, Liu, Zhang, Wang, Wang, et~al.]{wu2025verios}
Zheng Wu, Heyuan Huang, Xingyu Lou, Xiangmou Qu, Pengzhou Cheng, Zongru Wu, Weiwen Liu, Weinan Zhang, Jun Wang, Zhaoxiang Wang, et~al.
\newblock Verios: Query-driven proactive human-agent-gui interaction for trustworthy os agents.
\newblock \emph{arXiv preprint arXiv:2509.07553}, 2025.

\bibitem[Wu et~al.(2024)Wu, Wu, Xu, Wang, Sun, Jia, Cheng, Ding, Chen, Liang, et~al.]{wu2024atlas}
Zhiyong Wu, Zhenyu Wu, Fangzhi Xu, Yian Wang, Qiushi Sun, Chengyou Jia, Kanzhi Cheng, Zichen Ding, Liheng Chen, Paul~Pu Liang, et~al.
\newblock Os-atlas: A foundation action model for generalist gui agents.
\newblock \emph{arXiv preprint arXiv:2410.23218}, 2024.

\bibitem[Xu et~al.(2025{\natexlab{a}})Xu, Li, Yang, Zhang, Sun, Chow, Li, Song, Xu, Tong, Li, and Fei]{xu2025mixed}
Shilin Xu, Yanwei Li, Rui Yang, Tao Zhang, Yueyi Sun, Wei Chow, Linfeng Li, Hang Song, Qi~Xu, Yunhai Tong, Xiangtai Li, and Hao Fei.
\newblock Mixed-r1: Unified reward perspective for reasoning capability in multimodal large language models.
\newblock \emph{arXiv preprint arXiv:2505.24164}, 2025{\natexlab{a}}.

\bibitem[Xu et~al.(2025{\natexlab{b}})Xu, Wang, Wang, Lu, Xie, Saha, Sahoo, Yu, and Xiong]{xu2025aguvis}
Yiheng Xu, Zekun Wang, Junli Wang, Dunjie Lu, Tianbao Xie, Amrita Saha, Doyen Sahoo, Tao Yu, and Caiming Xiong.
\newblock Aguvis: Unified pure vision agents for autonomous {GUI} interaction.
\newblock In \emph{Forty-second International Conference on Machine Learning}, 2025{\natexlab{b}}.
\newblock URL \url{https://openreview.net/forum?id=PlihOwfx4r}.

\bibitem[Yao et~al.(2024)Yao, Yu, Zhang, Wang, Cui, Zhu, Cai, Li, Zhao, He, et~al.]{yao2024minicpm}
Yuan Yao, Tianyu Yu, Ao~Zhang, Chongyi Wang, Junbo Cui, Hongji Zhu, Tianchi Cai, Haoyu Li, Weilin Zhao, Zhihui He, et~al.
\newblock Minicpm-v: A gpt-4v level mllm on your phone.
\newblock \emph{arXiv preprint arXiv:2408.01800}, 2024.
\newblock \doi{10.48550/arXiv.2408.01800}.

\bibitem[Ye et~al.(2025)Ye, Zhang, Xu, Liu, Wang, Zhu, Zheng, Gao, Cao, Lu, et~al.]{ye2025mobile}
Jiabo Ye, Xi~Zhang, Haiyang Xu, Haowei Liu, Junyang Wang, Zhaoqing Zhu, Ziwei Zheng, Feiyu Gao, Junjie Cao, Zhengxi Lu, et~al.
\newblock Mobile-agent-v3: Foundamental agents for gui automation.
\newblock \emph{arXiv preprint arXiv:2508.15144}, 2025.

\bibitem[Zhang et~al.(2024{\natexlab{a}})Zhang, He, Qian, Li, Li, Qin, Kang, Ma, Liu, Lin, et~al.]{zhang2024large}
Chaoyun Zhang, Shilin He, Jiaxu Qian, Bowen Li, Liqun Li, Si~Qin, Yu~Kang, Minghua Ma, Guyue Liu, Qingwei Lin, et~al.
\newblock Large language model-brained gui agents: A survey.
\newblock \emph{arXiv preprint arXiv:2411.18279}, 2024{\natexlab{a}}.

\bibitem[Zhang et~al.(2025{\natexlab{a}})Zhang, Yang, Liu, Li, Han, Chen, Huang, Fu, and Yu]{zhang2025appagent}
Chi Zhang, Zhao Yang, Jiaxuan Liu, Yanda Li, Yucheng Han, Xin Chen, Zebiao Huang, Bin Fu, and Gang Yu.
\newblock Appagent: Multimodal agents as smartphone users.
\newblock In \emph{Proceedings of the 2025 CHI Conference on Human Factors in Computing Systems}, pp.\  1--20, 2025{\natexlab{a}}.

\bibitem[Zhang et~al.(2024{\natexlab{b}})Zhang, Wu, Teng, Liao, Xu, Xiao, Wei, and Tang]{zhang2024android}
Jiwen Zhang, Jihao Wu, Yihua Teng, Minghui Liao, Nuo Xu, Xiao Xiao, Zhongyu Wei, and Duyu Tang.
\newblock Android in the zoo: Chain-of-action-thought for gui agents.
\newblock \emph{arXiv preprint arXiv:2403.02713}, 2024{\natexlab{b}}.

\bibitem[Zhang et~al.(2024{\natexlab{c}})Zhang, Wang, Jia, Zheng, Yan, Gao, Li, and Xu]{zhang2024llamatouch}
Li~Zhang, Shihe Wang, Xianqing Jia, Zhihan Zheng, Yunhe Yan, Longxi Gao, Yuanchun Li, and Mengwei Xu.
\newblock Llamatouch: A faithful and scalable testbed for mobile ui automation task evaluation.
\newblock \emph{CoRR}, 2024{\natexlab{c}}.

\bibitem[Zhang et~al.(2025{\natexlab{b}})Zhang, Gao, and Xu]{zhang2025does}
Li~Zhang, Longxi Gao, and Mengwei Xu.
\newblock Does chain-of-thought reasoning help mobile gui agent? an empirical study.
\newblock \emph{arXiv preprint arXiv:2503.16788}, 2025{\natexlab{b}}.

\bibitem[Zhang et~al.(2025{\natexlab{c}})Zhang, Zhang, Fu, Wang, Yang, Du, Cui, Qin, Huang, Luo, and Luan]{zhang2025btl}
Shaojie Zhang, Ruoceng Zhang, Pei Fu, Shaokang Wang, Jiahui Yang, Xin Du, Shiqi Cui, Bin Qin, Ying Huang, Zhenbo Luo, and Jian Luan.
\newblock Btl-ui: Blink-think-link reasoning model for gui agent.
\newblock \emph{arXiv preprint arXiv:2509.15566}, 2025{\natexlab{c}}.
\newblock URL \url{https://arxiv.org/abs/2509.15566}.

\bibitem[Zhang et~al.(2025{\natexlab{d}})Zhang, Yu, and Yang]{zhang-etal-2025-attacking}
Yanzhe Zhang, Tao Yu, and Diyi Yang.
\newblock Attacking vision-language computer agents via pop-ups.
\newblock In Wanxiang Che, Joyce Nabende, Ekaterina Shutova, and Mohammad~Taher Pilehvar (eds.), \emph{Proceedings of the 63rd Annual Meeting of the Association for Computational Linguistics (Volume 1: Long Papers)}, pp.\  8387--8401, Vienna, Austria, July 2025{\natexlab{d}}. Association for Computational Linguistics.
\newblock ISBN 979-8-89176-251-0.
\newblock \doi{10.18653/v1/2025.acl-long.411}.
\newblock URL \url{https://aclanthology.org/2025.acl-long.411/}.

\bibitem[Zhang et~al.(2025{\natexlab{e}})Zhang, Liu, Tao, Chen, Fei, Che, and Qin]{zhang2025vitcot}
Yongheng Zhang, Xu~Liu, Ruihan Tao, Qiguang Chen, Hao Fei, Wanxiang Che, and Libo Qin.
\newblock Vitcot: Video-text interleaved chain-of-thought for boosting video understanding in large language models.
\newblock \emph{arXiv preprint arXiv:2507.09876}, 2025{\natexlab{e}}.

\bibitem[Zhang et~al.(2025{\natexlab{f}})Zhang, Lu, Fu, Huo, Yang, Wu, Si, Cong, Chen, Lin, et~al.]{zhang2025agentcpm}
Zhong Zhang, Yaxi Lu, Yikun Fu, Yupeng Huo, Shenzhi Yang, Yesai Wu, Han Si, Xin Cong, Haotian Chen, Yankai Lin, et~al.
\newblock Agentcpm-gui: Building mobile-use agents with reinforcement fine-tuning.
\newblock \emph{arXiv preprint arXiv:2506.01391}, 2025{\natexlab{f}}.

\bibitem[Zhang \& Zhang(2023)Zhang and Zhang]{zhang2023you}
Zhuosheng Zhang and Aston Zhang.
\newblock You only look at screens: Multimodal chain-of-action agents.
\newblock \emph{arXiv preprint arXiv:2309.11436}, 2023.

\bibitem[Zhao et~al.(2025)Zhao, Tan, Ma, Li, Jiang, Wang, Yang, and Liu]{zhao2025chain}
Chengshuai Zhao, Zhen Tan, Pingchuan Ma, Dawei Li, Bohan Jiang, Yancheng Wang, Yingzhen Yang, and Huan Liu.
\newblock Is chain-of-thought reasoning of llms a mirage? a data distribution lens.
\newblock \emph{arXiv preprint arXiv:2508.01191}, 2025.

\bibitem[Zheng et~al.(2024)Zheng, Gou, Kil, Sun, and Su]{zheng2024gpt}
Boyuan Zheng, Boyu Gou, Jihyung Kil, Huan Sun, and Yu~Su.
\newblock Gpt-4v (ision) is a generalist web agent, if grounded.
\newblock In \emph{International Conference on Machine Learning}, pp.\  61349--61385. PMLR, 2024.

\bibitem[Zhou et~al.(2025)Zhou, Dai, Wang, Zhou, Jia, and Xu]{zhou2025gui}
Yuqi Zhou, Sunhao Dai, Shuai Wang, Kaiwen Zhou, Qinglin Jia, and Jun Xu.
\newblock Gui-g1: Understanding r1-zero-like training for visual grounding in gui agents.
\newblock \emph{arXiv preprint arXiv:2505.15810}, 2025.

\end{thebibliography}
\bibliographystyle{iclr2026_conference}

\newpage
\appendix
\section{Appendix: Dataset Statistics and Sampling Strategy}\label{app:datasets}

In this appendix, we provide detailed statistics of the three evaluation datasets and describe the stratified sampling procedure used to construct the subset for human annotation.

\subsection{Action Type Distributions}

Table~\ref{tab:full-dist} summarizes the full action type distributions across AITZ, CAGUI, and AndroidControl. As can be seen, \texttt{CLICK} and \texttt{SCROLL} dominate most datasets, while other types such as \texttt{PRESS} or \texttt{LONG POINT} appear much less frequently.

\begin{table}[ht]
\centering
\small
\setlength{\tabcolsep}{5pt}
\begin{tabular}{lrrrrrr}
\toprule
\textbf{Action Type} &
\multicolumn{2}{c}{\textbf{AITZ}} &
\multicolumn{2}{c}{\textbf{CAGUI}} &
\multicolumn{2}{c}{\textbf{AndroidControl}} \\
\cmidrule(lr){2-3} \cmidrule(lr){4-5} \cmidrule(lr){6-7}
& Count & Ratio & Count & Ratio & Count & Ratio \\
\midrule
CLICK       & 2736 & 57.92\% & 3237 & 71.68\% & 5504 & 54.17\% \\
STOP        & 504  & 10.67\% & 600  & 13.29\% & 1680 & 16.53\% \\
SCROLL      & 601  & 12.72\% & 79   & 1.75\%  & 1297 & 12.76\% \\
INPUT       & 500  & 10.58\% & 574  & 12.71\% & 685  & 6.74\%  \\
NO\_ACTION  & 0    & 0.00\%  & 1    & 0.02\%  & 623  & 6.13\%  \\
PRESS       & 383  & 8.11\%  & 0    & 0.00\%  & 372  & 3.66\%  \\
LONG POINT  & 0    & 0.00\%  & 25   & 0.55\%  & 0    & 0.00\%  \\
\midrule
\textbf{Total} & 4724 & 100\% & 4516 & 100\% & 10161 & 100\% \\
\bottomrule
\end{tabular}
\caption{Full action type distribution across the three datasets.}
\label{tab:full-dist}
\end{table}

To obtain a balanced yet representative evaluation set, we applied a stratified sampling scheme. Table~\ref{tab:sampled-dist} shows the resulting distribution after sampling 200 instances from each dataset while preserving diversity across action types.

\begin{table}[ht]
\centering
\small
\setlength{\tabcolsep}{5pt}
\begin{tabular}{lrrrrrr}
\toprule
\textbf{Action Type} &
\multicolumn{2}{c}{\textbf{AITZ}} &
\multicolumn{2}{c}{\textbf{CAGUI}} &
\multicolumn{2}{c}{\textbf{AndroidControl}} \\
\cmidrule(lr){2-3} \cmidrule(lr){4-5} \cmidrule(lr){6-7}
& Count & Ratio & Count & Ratio & Count & Ratio \\
\midrule
CLICK       & 106 & 53.00\% & 131 & 65.50\% & 106 & 53.00\% \\
STOP        & 24  & 12.00\% & 28  & 14.00\% & 36  & 18.00\% \\
SCROLL      & 27  & 13.50\% & 8   & 4.00\%  & 29  & 14.50\% \\
INPUT       & 24  & 12.00\% & 27  & 13.50\% & 17  & 8.50\%  \\
PRESS       & 19  & 9.50\%  & 0   & 0.00\%  & 12  & 6.00\%  \\
LONG POINT  & 0   & 0.00\%  & 6   & 3.00\%  & 0   & 0.00\%  \\
\midrule
\textbf{Total} & 200 & 100\% & 200 & 100\% & 200 & 100\% \\
\bottomrule
\end{tabular}
\caption{Action type distribution after stratified sampling (200 samples per dataset).}
\label{tab:sampled-dist}
\end{table}

\subsection{Sampling Method}

We adopted a stratified sampling procedure with minimum allocation and paired projection. 
For each dataset with stratum counts $\{n_c\}$ and target size $N$, we first allocate
\[
m_c = \min(k,n_c),\quad M=\sum_c m_c,\quad R=N-M,
\]
then distribute the remainder proportionally using the largest remainder method:
\[
t_c = \min\!\bigl(n_c,\, m_c+\lfloor R\cdot n_c/\sum_j n_j \rfloor+\delta_c\bigr),\qquad \sum_c t_c=N.
\]
On a baseline model, we sample per stratum to form a key list $\texttt{key\_list}=\{(\texttt{episode\_id},\texttt{step\_id})\}$,
and then project the same keys onto other models to guarantee alignment.

\begin{lstlisting}[language=Python, caption={Stratified sampling with paired projection}]
# Input: counts {n_c}, total N, minimum k, dataset D
m_c = min(k, n_c)           # per stratum minimum
M = sum(m_c for c in C)
R = N - M
q_c = R * n_c / sum(n_j)    # proportional share
a_c = floor(q_c)
L = R - sum(a_c)
delta_c = distribute_L_by_largest_remainder(q_c - a_c)
t_c = min(n_c, m_c + a_c + delta_c)

# Draw without replacement to build key_list
key_list = sample_per_stratum(D, t_c)
# For other models, filter by key_list
\end{lstlisting}

\section{Appendix: Models Deployment}\label{app:models}
\paragraph{AgentCPM-GUI~\citep{zhang2025agentcpm}.}
This model family adopts a single-turn format without history:
\begin{equation}
P(c_n, a_n \mid I, o_n),
\end{equation}
which is equivalent to setting $H_n = \varnothing$.

\begin{tcolorbox}[breakable, colback=black!5!white,colframe=black!75!black,title=AgentCPM-GUI Data Examples]
\tcbsubtitle{System Prompt}
\begin{lstlisting}
# Role
You are an intelligent agent familiar with Android touchscreen GUI operations.  
Based on the user's query, you will analyze the GUI elements and layout of the current interface, and generate the corresponding operation.  

# Task
Given the user's query and the current screen screenshot, output the next step operation.  

# Rule
- Output must be in compact JSON format.  
- The operation must follow the Schema constraints.  

# Schema
```json
{
  "type": "object",
  "description": "Execute an operation and decide the current task status",
  "additionalProperties": false,
  "properties": {
    "thought": { 
      "type": "string", 
      "description": "Reasoning process of the agent" 
    },
    "POINT": { 
      "$ref": "#/$defs/Location", 
      "description": "Tap at the specified position on the screen" 
    },
    "to": {
      "description": "Movement, combined gesture parameters",
      "oneOf": [
        {
          "enum": [ "up", "down", "left", "right" ],
          "description": "From the current point (POINT), perform a swipe gesture in one of the four directions"
        },
        {
          "$ref": "#/$defs/Location",
          "description": "Move to a specific position"
        }
      ]
    },
    "duration": {
      "type": "integer",
      "description": "Execution time or waiting time in milliseconds",
      "minimum": 0,
      "default": 200
    },
    "PRESS": {
      "type": "string",
      "description": "Trigger special keys. HOME = go to home screen, BACK = back button, ENTER = enter key",
      "enum": [ "HOME", "BACK", "ENTER" ]
    },
    "TYPE": { 
      "type": "string", 
      "description": "Input text" 
    },
    "STATUS": {
      "type": "string",
      "description": "Current task status. Special cases: satisfied = no action required; impossible = task cannot be completed; interrupt = task interrupted; need_feedback = user feedback required",
      "enum": [ "continue", "finish", "satisfied", "impossible", "interrupt", "need_feedback" ],
      "default": "continue"
    }
  },
  "$defs": {
    "Location": {
      "type": "array",
      "description": "Coordinates relative to the top-left corner of the screen, scaled between 0-1000 by width and height. First element = x, second element = y",
      "items": { "type": "integer", "minimum": 0, "maximum": 1000 },
      "minItems": 2,
      "maxItems": 2
    }
  }
}
\end{lstlisting}
\tcbsubtitle{User}
\begin{lstlisting}
<Question>[query]</Question>\nCurrent screen screenshot:
[current_screenshot]
\end{lstlisting}
\tcbsubtitle{Assistant}
[thought\_and\_action]
\end{tcolorbox}

\paragraph{UI-TARS~\citep{qin2025ui}.}
This model family organizes the task as a multi-turn dialogue, retaining at most the last $N$ interaction triples in the context. Specifically,
\begin{equation}
H_n = \{(o_j, c_j, a_j)\}_{j=\max(1,\,n-N)}^{\,n-1},
\end{equation}
where $N=4$ in our setting. The model predicts
\begin{equation}
P(c_n, a_n \mid I, H_n, o_n).
\end{equation}

\begin{tcolorbox}[breakable, colback=black!5!white,colframe=black!75!black,title=UI-TARS Data Example]
\tcbsubtitle{System Message}
You are a helpful assistant.
\tcbsubtitle{User}
\begin{lstlisting}
You are a GUI agent. You are given a task and your action history, with screenshots. You need to perform the next action to complete the task. 
## Output Format
```
Thought: ...
Action: ...
```
## Action Space

click(point='<point>x1 y1</point>')
long_press(point='<point>x1 y1</point>')
type(content='') #If you want to submit your input, use "\\n" at the end of `content`.
scroll(point='<point>x1 y1</point>', direction='down or up or right or left')
press_home()
press_back()
finished(content='xxx') # Use escape characters \\', \\", and \\n in content part to ensure we can parse the content in normal python string format.


## Note
- Use {language} in `Thought` part.
- Write a small plan and finally summarize your next action (with its target element) in one sentence in `Thought` part.

## User Instruction
{instruction}
\end{lstlisting}
\tcbsubtitle{User}
[history\_screenshot]
\tcbsubtitle{Assistant}
[history\_thought\_and\_action]
\tcbsubtitle{User}
[current\_screenshot]
\tcbsubtitle{Assistant}
[thought\_and\_action]
\label{eg_uitars}
\end{tcolorbox}

\paragraph{GUI-Owl~\citep{ye2025mobile}.}
This model family also uses a single-turn format but supplements the input with a compressed history representation $\tilde H_n$:
\begin{equation}
P(c_n, a_n \mid I, \tilde H_n, o_n),
\end{equation}
where $\tilde H_n = \mathrm{Compress}(H_n)$ denotes a textual summary of past interactions.

\begin{tcolorbox}[breakable, colback=black!5!white,colframe=black!75!black,title=GUI-Owl Data Example]
\tcbsubtitle{System Message}
\begin{lstlisting}
You are a helpful assistant.

# Tools

You may call one or more functions to assist with the user query.

You are provided with function signatures within <tools></tools> XML tags:
<tools>
{"type": "function", "function": {"name_for_human": "mobile\_use", "name": "mobile\_use", "description": "Use a touchscreen to interact with a mobile device, and take screenshots.
* This is an interface to a mobile device with touchscreen. You can perform actions like clicking, typing, swiping, etc.
* Some applications may take time to start or process actions, so you may need to wait and take successive screenshots to see the results of your actions.
* The screen's resolution is {width}x{height}.
* Make sure to click any buttons, links, icons, etc with the cursor tip in the center of the element. Don't click boxes on their edges unless asked.", "parameters": {"properties": {"action": {"description": "The action to perform. The available actions are:
* `key`: Perform a key event on the mobile device.
    - This supports adb's `keyevent` syntax.
    - Examples: \"volume\_up\", \"volume\_down\", \"power\", \"camera\", \"clear\".
* `click`: Click the point on the screen with coordinate (x, y).
* `long\_press`: Press the point on the screen with coordinate (x, y) for specified seconds.
* `swipe`: Swipe from the starting point with coordinate (x, y) to the end point with coordinates2 (x2, y2).
* `type`: Input the specified text into the activated input box.
* `system\_button`: Press the system button.
* `open`: Open an app on the device.
* `wait`: Wait specified seconds for the change to happen.
* `terminate`: Terminate the current task and report its completion status.", "enum": ["key", "click", "long\_press", "swipe", "type", "system\_button", "open", "wait", "terminate"], "type": "string"}, "coordinate": {"description": "(x, y): The x (pixels from the left edge) and y (pixels from the top edge) coordinates to move the mouse to. Required only by `action=click`, `action=long\_press`, and `action=swipe`.", "type": "array"}, "coordinate2": {"description": "(x, y): The x (pixels from the left edge) and y (pixels from the top edge) coordinates to move the mouse to. Required only by `action=swipe`.", "type": "array"}, "text": {"description": "Required only by `action=key`, `action=type`, and `action=open`.", "type": "string"}, "time": {"description": "The seconds to wait. Required only by `action=long\_press` and `action=wait`.", "type": "number"}, "button": {"description": "Back means returning to the previous interface, Home means returning to the desktop, Menu means opening the application background menu, and Enter means pressing the enter. Required only by `action=system\_button`", "enum": ["Back", "Home", "Menu", "Enter"], "type": "string"}, "status": {"description": "The status of the task. Required only by `action=terminate`.", "type": "string", "enum": ["success", "failure"]}}, "required": ["action"], "type": "object"}, "args\_format": "Format the arguments as a JSON object."}}
</tools>

For each function call, return a json object with function name and arguments within <tool\_call></tool\_call> XML tags:
<tool_call>
{"name": <function-name>, "arguments": <args-json-object>}
</tool_call>
\end{lstlisting}
\tcbsubtitle{User}
\begin{lstlisting}
The user query:  [user_request]
Task progress (You have done the following operation on the current device): [history_actions]
Before answering, explain your reasoning step-by-step in <thinking></thinking> tags, and insert them before the <tool_call></tool_call> XML tags.
After answering, summarize your action in <conclusion></conclusion> tags, and insert them after the <tool_call></tool_call> XML tags.

[current_screenshot]
\end{lstlisting}
\tcbsubtitle{Assistant}
[thought\_and\_action]
\label{eg_qwen}
\end{tcolorbox}

\section{Appendix: Evaluation Procedures}\label{app:evaluation}
To ensure reproducibility of our results, we release supplementary code together with detailed 
instructions in the repository. The main evaluation procedures are as follows:

\begin{enumerate}
    \item \textbf{Environment setup.} All experiments were conducted on Linux with Python~3.10/3.11 
    and CUDA-enabled GPUs. The repository provides a \texttt{requirements.txt} file to reproduce 
    the exact software environment.
    \item \textbf{Model preparation.} Pre-trained checkpoints of AgentCPM-GUI, UI-TARS, and 
    GUI-Owl are downloaded following the official instructions (see \texttt{model/README.md}). 
    Local paths are configured as required by the inference scripts.
    \item \textbf{Dataset preparation.} Evaluation datasets (AITZ, CAGUI, AndroidControl) are 
    publicly available. We provide preprocessing scripts and instructions in 
    \texttt{eval/eval\_data/README.md}, which reproduce the splits used in our experiments.
    \item \textbf{Inference and EM evaluation.} For each model–dataset pair, we run inference 
    scripts (\texttt{run\_predict\_*.py}) to generate action predictions, followed by 
    \texttt{run\_eval\_agent.py} to compute Exact Match (EM) scores.
    \item \textbf{GTA evaluation.} To evaluate reasoning correctness, we provide scripts to extract CoT traces and apply the GTA evaluator 
    (\texttt{eval/README\_COT.md}). The evaluation enforces strict type and parameter matching 
    consistent with the EM metric.
    \item \textbf{Manual annotation (optional).} For RQ1, we use a lightweight annotation interface 
    (\texttt{cot\_eval/annot\_ui\_min.py}) to construct a human-annotated reference set. 
    Disagreements are filtered, and only consensus labels are retained.
    \item \textbf{Analysis and visualization.} Quadrant analysis and diagnostic plots are 
    generated with scripts in \texttt{cot\_eval/}, reproducing all figures reported in the main text.
\end{enumerate}

This procedure allows independent researchers to replicate all experiments reported in the paper, 
from model inference to reasoning–execution gap analysis.

\section{Appendix: Additional Results}\label{app:additional}
In this appendix, we provide supplementary quantitative results to complement the main text. 
Table~\ref{tab:reliability} reports the reliability of the GTA Evaluator, showing that it achieves 
consistently high accuracy across datasets and models. Table~\ref{tab:benchmarks} presents 
benchmark results on AITZ, CAGUI, and AndroidControl, offering a more detailed comparison 
of reasoning and execution metrics (EM, GTA, EG, and RG). Together, these results further 
validate the robustness of our evaluation framework and highlight distinct performance patterns 
across model families and datasets.
\begin{table}[h]
\centering
\small
\setlength{\tabcolsep}{3.6pt}
\begin{tabular}{lccccccccc}
\toprule
\multirow{2}{*}{Dataset} & \multicolumn{3}{c}{AgentCPM-GUI-8B} & \multicolumn{3}{c}{UI-TARS-1.5-7B} & \multicolumn{3}{c}{GUI-Owl-7B} \\
\cmidrule(lr){2-4} \cmidrule(lr){5-7} \cmidrule(lr){8-10}
 & AITZ & CAGUI & AC & AITZ & CAGUI & AC & AITZ & CAGUI & AC \\
\midrule
Valid samples      & 159 & 169 & 135 & 153 & 168 & 155 & 150 & 180 & 162 \\
GTA Accuracy (\%) & 77.36 & 82.84 & 88.89 & 86.93 & 89.88 & 92.90 & 84.00 & 78.89 & 93.83 \\
\bottomrule
\end{tabular}
\caption{Reliability of the GTA Evaluator. Overall, the evaluator achieves consistently high accuracy, with similar performance across models. Accuracy peaks on AndroidControl (AC), while results on CAGUI and AITZ are slightly lower.} 
\label{tab:reliability}
\end{table}

\begin{table}[h]
\centering
\small
\setlength{\tabcolsep}{3.4pt}
\begin{tabular}{lcccc cccc cccc}
\toprule
\multirow{2}{*}{\textbf{Model}}
& \multicolumn{4}{c}{\textbf{AITZ} (\%)} 
& \multicolumn{4}{c}{\textbf{CAGUI} (\%)}
& \multicolumn{4}{c}{\textbf{AndroidControl} (\%)} \\
\cmidrule(lr){2-5} \cmidrule(lr){6-9} \cmidrule(lr){10-13}
 & EM & GTA & EG & RG & EM & GTA & EG & RG & EM & GTA & EG & RG \\
\midrule
AgentCPM-GUI-8B & \textbf{76.29} & \textbf{70.43} & \textbf{3.51}  & 9.38 & \textbf{91.21} & \textbf{85.35} & \textbf{2.28}  & 8.13 & 69.17 & 74.45 & 8.68  & 3.41 \\
UI-TARS-7B-SFT  & 66.69 & 64.98 & 7.73 & 9.44 & 71.36 & 77.86 & 9.08 & 2.57 & 74.71 & \textbf{81.44} & 9.18  & \textbf{2.45} \\
UI-TARS-7B-DPO  & 65.95 & 63.86 & 8.48 & 10.57 & 70.60 & 78.52 & 10.35 & 2.44 & 73.87 & 80.16 & 9.15  & 2.85 \\
UI-TARS-1.5-7B  & 58.94 & 66.09 & 11.85 & \textbf{4.69} & 68.42 & 81.68 & 15.32 & \textbf{2.06} & \textbf{76.10} & 76.22 & \textbf{5.53}  & 5.41 \\
GUI-Owl-7B      & 61.01 & 66.17 & 10.29 & 5.12 & 61.00 & 79.87 & 21.75 & 2.88 & 65.22 & 72.79 & 12.31 & 4.74 \\
GUI-Owl-32B     & 59.65 & 66.09 & 12.19 & 5.76 & 65.88 & 81.53 & 18.29 & 2.64 & 69.89 & 74.15 & 10.43 & 6.18 \\
\bottomrule
\end{tabular}
\caption{Benchmark results on AITZ, CAGUI, and AndroidControl datasets. Best results are marked in bold face. Overall, AgentCPM-GUI shows leading performance on AITZ and CAGUI, while UI-TARS achieves the strongest results on AndroidControl.}
\label{tab:benchmarks}
\end{table}



\section{Appendix: The Use of Large Language Models}

In preparing this paper, we used ChatGPT~5 to aid or polish the writing, particularly for grammar 
and stylistic refinement. The authors take full responsibility for all the content presented under 
their names, including any parts where LLMs were involved. The use of LLMs did not affect the 
design, execution, or interpretation of the research itself. 

\section{Appendix: Case Study}
In the following case studies, we visualize four representative outcomes of reasoning–execution
alignment using \textbf{UI-TARS-1.5-7B}. Each figure shows the mobile screenshot annotated with
two markers: \textcolor{green}{green dots} denote the ground-truth click locations, while
\textcolor{blue}{blue dots} indicate the model’s predicted clicks. The examples cover all four
quadrants of our diagnostic framework: Both Right (Ideal), Both Wrong, Execution Gap (EG),
and Reasoning Gap (RG).

\label{appendix:case}
\begin{figure}[t]
    \centering
    \begin{minipage}{0.48\linewidth}
        \small
        \textbf{Case: Both Right (Ideal)} \\[3pt]
        \textbf{Instruction:} Add Lenovo ThinkPad to the cart on Target. \\[3pt]
        \textbf{CoT:} I noticed that the search results page has loaded, and the first item displayed is the Lenovo ThinkPad T570 laptop. To proceed with the task of adding it to the cart, I need to click on the product image or title to access the product details page. This will allow me to see more information about the product and the option to add it to the cart. \\[3pt]
        \textbf{Prediction:} \texttt{CLICK(1259, 414)} \\[3pt]
        \textbf{Ground Truth:} \texttt{CLICK(1262, 464)} \\[3pt]
        \textbf{Scores:} $\mathrm{EM}=1$, $\mathrm{GTA}=1$
    \end{minipage}
    \hfill
    \begin{minipage}{0.48\linewidth}
        \centering
        \includegraphics[width=\linewidth]{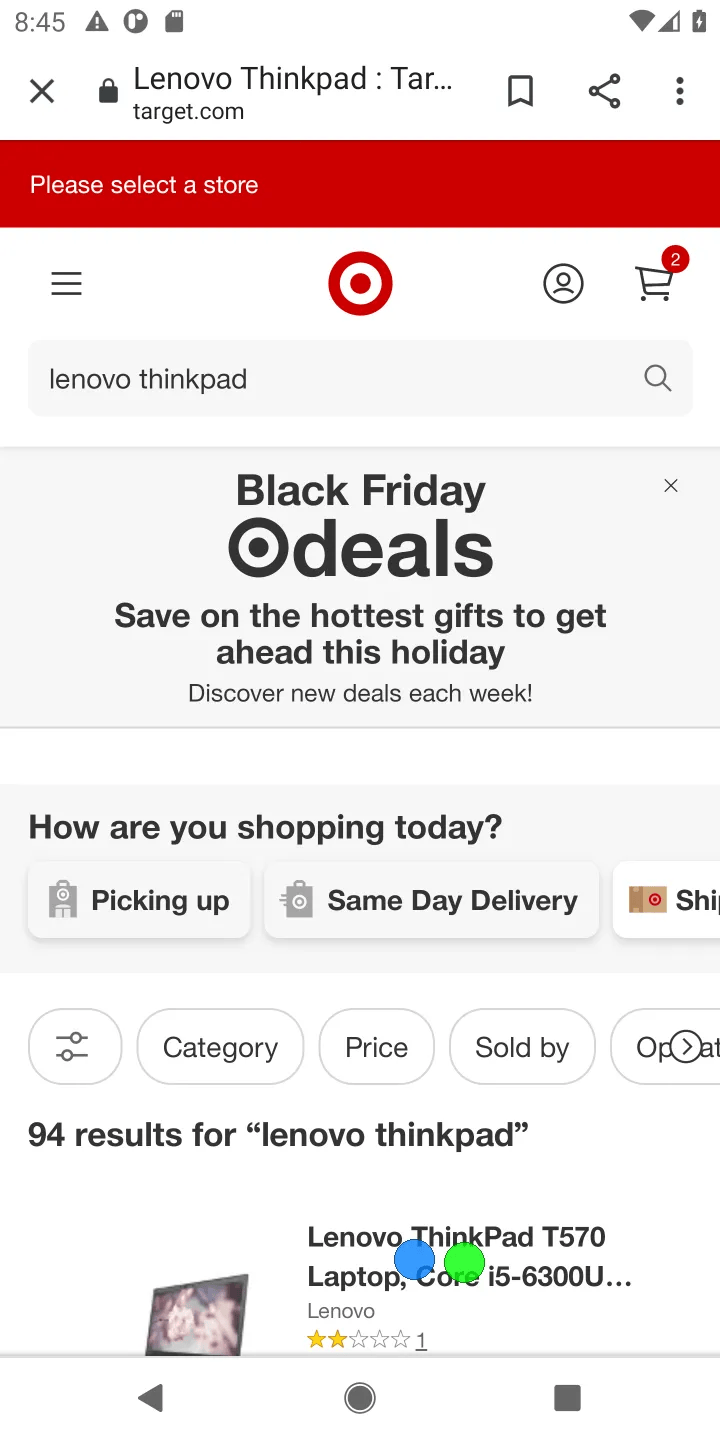}
    \end{minipage}
    \caption{Case study of a Both Right (Ideal) example.}
    \label{fig:case_bothright}
\end{figure}

\begin{figure}[t]
    \centering
    \begin{minipage}{0.48\linewidth}
        \small
        \textbf{Case: Both Wrong} \\[3pt]
        \textbf{Instruction:} Check the best rated table lamp on Crate and Barrel. \\[3pt]
        \textbf{CoT:} I noticed that there's a pop-up at the bottom of the page offering a 10\% discount, which is going to interfere with my ability to browse the products. I need to close it first, so I'll click the close button in the upper right corner. This way, I can focus on finding the best-rated table lamps. \\[3pt]
        \textbf{Prediction:} \texttt{CLICK(864, 488)} \\[3pt]
        \textbf{Ground Truth:} \texttt{CLICK(596, 56)} \\[3pt]
        \textbf{Scores:} $\mathrm{EM}=0$, $\mathrm{GTA}=0$
    \end{minipage}
    \hfill
    \begin{minipage}{0.48\linewidth}
        \centering
        \includegraphics[width=\linewidth]{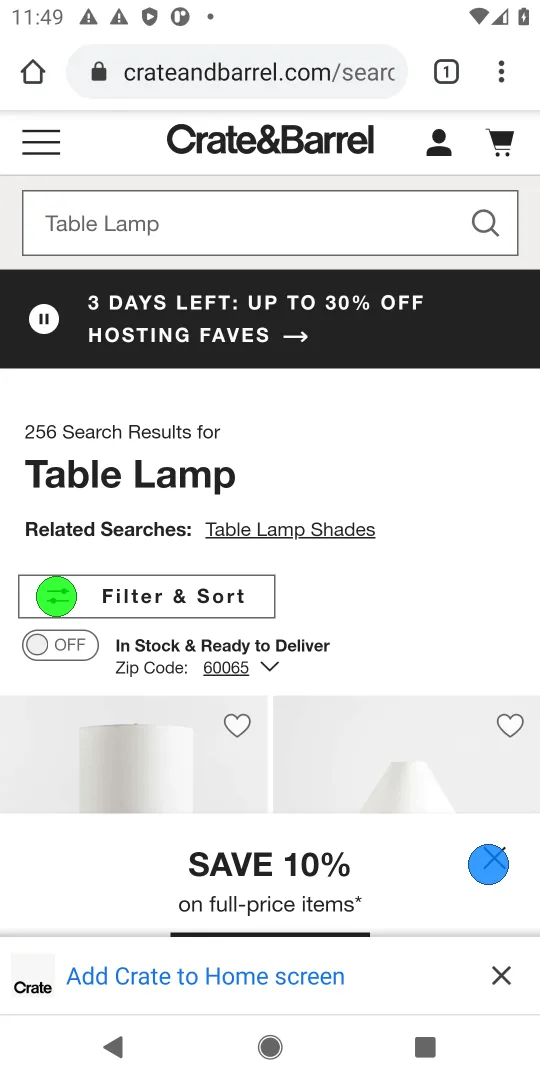}
    \end{minipage}
    \caption{Case study of a Both Wrong example.}
    \label{fig:case_bothwrong_crate}
\end{figure}

\begin{figure}[t]
    \centering
    \begin{minipage}{0.48\linewidth}
        \small
        \textbf{Case: Execution Gap (EG)} \\[3pt]
        \textbf{Instruction:} Go to internet settings. \\[3pt]
        \textbf{CoT:} I noticed that the current page is the "Hotspot \& tethering" settings interface, which is not the internet settings we are looking for. Let me click the back button in the upper left corner to return to the previous menu and continue searching for the correct settings option. \\[3pt]
        \textbf{Prediction:} \texttt{STOP} (terminate task) \\[3pt]
        \textbf{Ground Truth:} \texttt{CLICK(82, 38)} \\[3pt]
        \textbf{Scores:} $\mathrm{EM}=0$, $\mathrm{GTA}=1$
    \end{minipage}
    \hfill
    \begin{minipage}{0.48\linewidth}
        \centering
        \includegraphics[width=\linewidth]{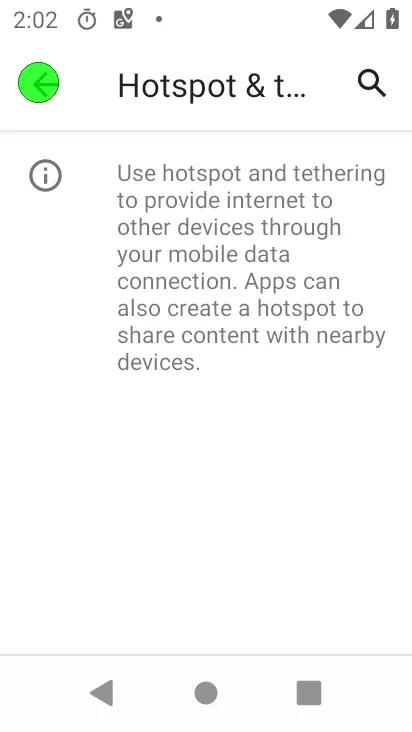}
    \end{minipage}
    \caption{Case study of an Execution Gap example.}
    \label{fig:case_eg_internet}
\end{figure}

\begin{figure}[t]
    \centering
    \begin{minipage}{0.48\linewidth}
        \small
        \textbf{Case: Reasoning Gap (RG)} \\[3pt]
        \textbf{Instruction:} Since I am traveling, I would like to listen to news podcasts on the New York Times app. \\[3pt]
        \textbf{CoT:} I'm currently on the health section page of the New York Times app, and it seems that this isn't the right place to find podcasts. I need to go back to the homepage to look for the podcasts section. Let me click the back button in the top left corner to return to the main page and continue my search. \\[3pt]
        \textbf{Prediction:} \texttt{PRESS(back)} \\[3pt]
        \textbf{Ground Truth:} \texttt{PRESS(back)} \\[3pt]
        \textbf{Scores:} $\mathrm{EM}=1$, $\mathrm{GTA}=0$
    \end{minipage}
    \hfill
    \begin{minipage}{0.48\linewidth}
        \centering
        \includegraphics[width=\linewidth]{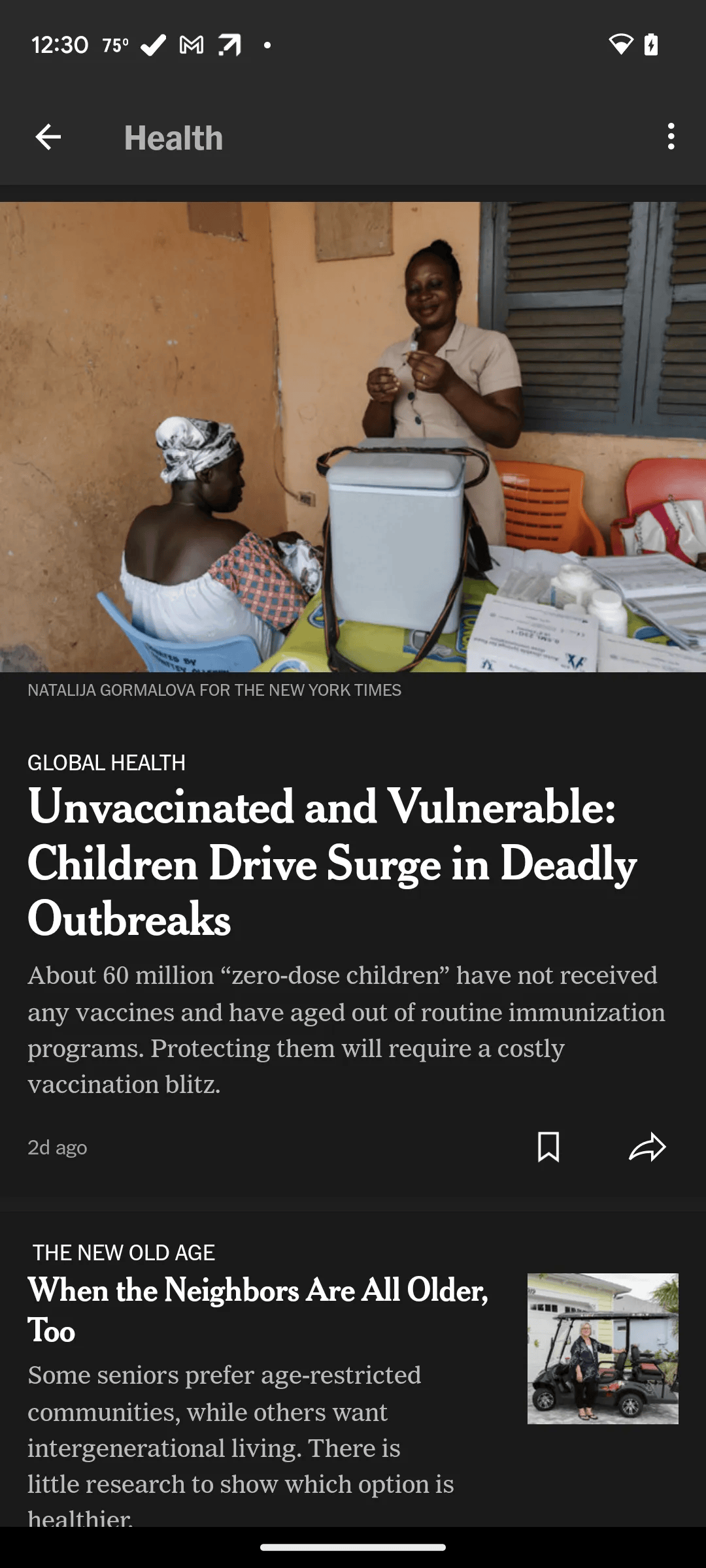}
    \end{minipage}
    \caption{Case study of a Reasoning Gap example.}
    \label{fig:case_rg}
\end{figure}

\end{document}